%% file: main.tex
\documentclass{article} 
\usepackage{iclr2027_conference,times}

\input{math_commands.tex}

\usepackage{graphicx}
\usepackage{hyperref}
\usepackage{url}
\usepackage{booktabs} 
\usepackage{multirow}
\usepackage{subcaption}
\usepackage{xcolor}
\newcommand{\cmidruletrim}[2]{\cmidrule(#1){#2}}

\title{Reuse or Relearn? A Spectral View of\\ Earth Observation Foundation Models}

\author{
Mehmet Ozgur Turkoglu\\
Agroscope, EO of Agroecosystems\\
\texttt{moturkoglu@gmail.com}
\And
Valerio Marsocci \\
ESA, $\Phi$-lab\\
\And
Dominik J. M{\"u}hlematter \\
ETH Zurich \\
\AND
Dominik Senti \\
ETH Zurich \&\\
Agroscope, EO of Agroecosystems \\
\And
Konrad Schindler \\
ETH Zurich \\
\And
Helge Aasen \\
Agroscope, EO of Agroecosystems \\
}

\iclrfinalcopy 
\begin{document}

\maketitle

\begin{abstract}

Foundation models are rarely used as generic, frozen feature extractors; instead, they are fine-tuned for the target downstream application. This practice is particularly prevalent in Earth observation (EO), and it raises a question that downstream accuracy alone cannot answer: does fine-tuning reuse the pretrained representation, or does it relearn a new one? We study this with spectral diagnostics that compare a model before and after adaptation, quantifying how well its dominant singular subspaces are preserved, how broadly the weight update is distributed, and how large it is. Using natural image models such as CLIP and DINO as a reference, we find that, under the evaluated fine-tuning settings, EO models undergo far larger, higher-rank updates and retain much less of their pretrained structure, so their downstream performance is often obtained with substantial changes to the pretrained weight structure. More broadly, foundation models, and EO foundation models in particular, should be assessed not only by benchmark accuracy, but also by how reusable their pretrained representation is.

\end{abstract}

\input{sections/01_intro}

\input{sections/03_method}

\input{sections/04_experiment}
\input{sections/06_discussion}

\input{sections/02_RW}

\input{sections/05_conclusion}

\section*{AI Use Statement}
We used generative AI tools as writing aids to improve the language, clarity, grammar, and readability of the manuscript. Generative AI tools were \emph{not} used for substantive research tasks, including developing the methodology, formulating hypotheses or mathematical claims, conducting experiments, analyzing data, or interpreting results. They were also used to assist in preparing the public code release. All AI-assisted text and code were reviewed by the authors to ensure correctness and consistency with the intended content. The authors take full responsibility for the final manuscript and released
code.

\section*{Ethics Statement}
This work studies existing, public Earth observation foundation models and does not involve human subjects or real-world deployment. We note that EO models can have dual-use applications and may reflect geographic or dataset biases.

\section*{Reproducibility Statement}
We provide the definitions and implementation details of all spectral diagnostics in Section~\ref{sec:method} and Appendix~\ref{app:spectral_details}. The experimental setup, including dataset splits, preprocessing, model configurations, adaptation methods, trainable parameter counts, optimization settings, and random seeds, is documented in Section~\ref{sec:exp} and Appendix~\ref{training_details}. We also provide the code required to reproduce experiments.

\bibliography{iclr2027_conference}
\bibliographystyle{iclr2027_conference}

\clearpage
\appendix
\input{appendix}

%

\end{document}

%% file: math_commands.tex
\usepackage{amsmath,amsfonts,bm}

\def\eqref#1{equation~\ref{#1}}

\def\1{\bm{1}}

\DeclareMathAlphabet{\mathsfit}{\encodingdefault}{\sfdefault}{m}{sl}
\SetMathAlphabet{\mathsfit}{bold}{\encodingdefault}{\sfdefault}{bx}{n}



%% file: sections/01_intro.tex
\section{Introduction}\label{sec:intro}
Foundation models aim to learn representations that can be reused across downstream tasks~\citep{bommasani2021foundation}. This promise has motivated large-scale pretraining in Earth observation (EO), with models intended to transfer across sensors, regions, and applications~\citep{cong2022satmae,fuller2023croma,szwarcman2025prithvi}. However, strong performance after fine-tuning does not reveal how much of the pretrained structure is retained or how it contributes to solving the downstream task. Head-only evaluation also provides an incomplete picture. A linear head measures how readily the task can be solved from frozen features, which may favor representations already aligned with that task without capturing their usefulness after adaptation. 
%
%
More expressive decoders, such as UPerNet \citep{xiao2018unified}, introduce substantial task-specific learning capacity, making it harder to attribute strong performance specifically to the pretrained features. The same applies to pre-computed Earth embeddings~\citep{brown2025alphaearth, feng2026tessera}, which are often paired with sizeable task heads and can still be outperformed by end-to-end fine-tuning~\citep{herzog2026olmoearth}. Understanding reuse therefore requires examining both the benefits of pretraining and the changes through which the pretrained model adapts to a new task.

Work in AI interpretability provides a basis for this analysis. Learned features have been associated with linear directions and subspaces, including interpretable singular directions in transformer weights~\citep{elhage2022superposition,millidge2022svd}, while spectral analyses indicate that information is distributed across singular modes~\citep{staats2025small}. Singular-value fine-tuning builds on this perspective by adjusting the strength of pretrained modes while preserving their directions~\citep{sun2022svf,turkoglu2026sve}. These findings motivate studying whether pretrained singular subspaces remain useful during adaptation.

  We combine established spectral measures into a diagnostic toolkit that characterizes adaptation along three complementary dimensions. The \emph{Spectral Stability Index} (SSI) measures preservation of dominant pretrained subspaces, the \emph{Effective Rank of Adaptation} (ERA) measures how many directions contribute to the update, and relative weight change measures its magnitude. Comparisons across pretrained model generations and downstream task requirements ground their interpretation. Together with downstream performance, these diagnostics provide a structural view of representation reuse.

We apply this framework to full fine-tuning across five EO benchmarks and compare downstream performance with parameter-efficient alternatives: singular-value fine-tuning (SVF) and LoRA~\citep{hu2022lora}. Experiments with three computer vision (CV) backbones provide a reference for adaptation and reuse. To assess which benchmarks reveal the value of pretraining, we compare pretrained models with architecture-matched, randomly initialized counterparts. 
We also provide additional analyses, including experiments on cross-domain and cross-modal transfer.
Our contributions are:
\begin{enumerate}

    \item \textbf{A spectral perspective on the usefulness and reusability of pretraining.}
    We combine established spectral measures into a diagnostic toolkit and
    empirically ground it for examining how pretrained weight structure
    supports downstream adaptation, extending evaluation beyond task-level
    performance.

    \item \textbf{Distinct adaptation behaviors in EO and CV foundation models.}
    We show that the evaluated EO models generally undergo larger updates with higher effective ranks and weaker subspace preservation, suggesting less mature representation reuse than in the CV reference. Our analyses also identify cases where strong performance requires comparatively limited changes, highlighting opportunities for more reusable pretraining.
\end{enumerate}

\begin{figure}[t]
    \centering
    \includegraphics[width=\linewidth]{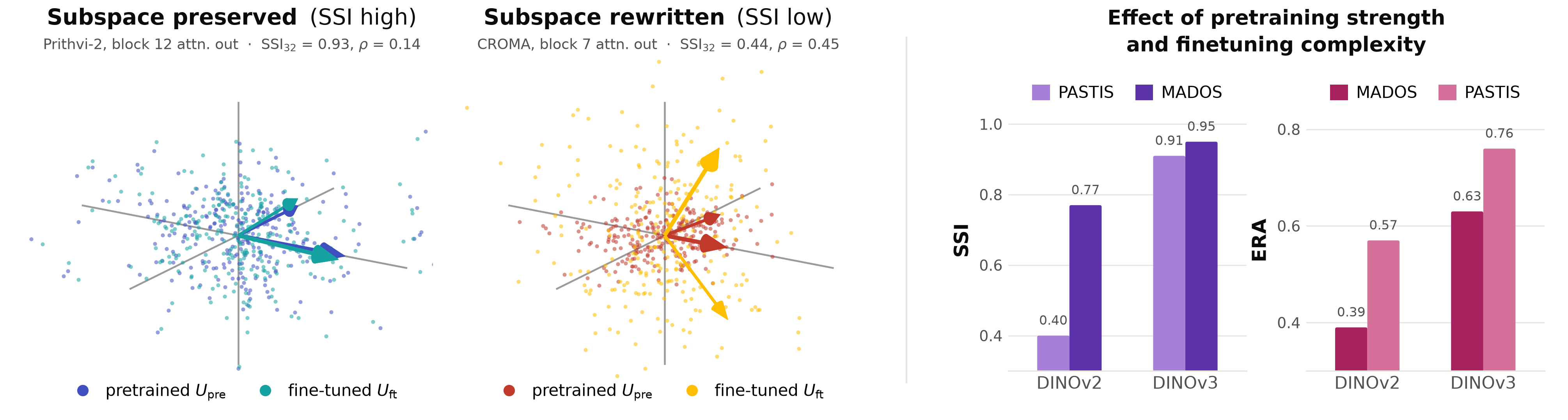}
\caption{
\textbf{Left:} One attention output projection $W \in \mathbb{R}^{n \times n}$ per panel, pretrained and after full fine-tuning on MADOS (Prithvi-2 block 12, $n = 768$ and CROMA block 7, $n = 1024$). Points are 260 sampled columns of $W$, arrows are the two leading left singular vectors, scaled by their singular values ($U_2 \Sigma_2$). Both are projected onto the 3-D subspace that best fits the four arrows. SSI$_k$ measures how well the pretrained and fine-tuned top-$k$ subspaces align, independently of singular-value magnitude.
\textbf{Right:} DINOv3 exhibits higher SSI than DINOv2 on both datasets, suggesting that stronger pretraining is associated with greater preservation of pretrained subspaces during fine-tuning. Both models exhibit higher ERA on PASTIS than on MADOS, suggesting that adaptation to the more complex task involves a broader range of update directions.}
\label{fig:ssi_era_interpretation}
\end{figure}

%% file: sections/03_method.tex
\section{Spectral Diagnostics of Fine-Tuning}
\label{sec:method}

We study how fine-tuning modifies a pretrained encoder in weight space.
For a pretrained weight matrix
$W_{\mathrm{pre}} \in \mathbb{R}^{m \times n}$ and its fine-tuned
counterpart $W_{\mathrm{ft}}$, we define the update
\begin{equation}
    \Delta W = W_{\mathrm{ft}} - W_{\mathrm{pre}}.
\end{equation}
We characterize this update along three complementary dimensions:
\emph{how much} the weights change, \emph{whether} their dominant
pretrained subspaces are preserved, and \emph{how many directions}
participate in the update. These are measured by relative weight
change, the Spectral Stability Index (SSI), and the Effective Rank of
Adaptation (ERA), respectively.

\subsection{Relative Weight Change}

We first measure the magnitude of adaptation relative to the scale of
the pretrained weights:
\begin{equation}
    \rho =
    \frac{\|\Delta W\|_F}{\|W_{\mathrm{pre}}\|_F}.
    \label{eq:relative_change}
\end{equation}
The relative weight change $\rho$ quantifies how strongly a matrix is
modified during fine-tuning. Small values indicate that the fine-tuned
weights remain close to their pretrained initialization, whereas larger
values indicate more substantial changes.

\subsection{Spectral Stability Index (SSI)}

Relative weight change measures the size of an update, but not whether
fine-tuning preserves the dominant directions of the pretrained
weights. To capture this, we compare the leading singular subspaces
before and after fine-tuning. We write the singular value decompositions as
\begin{equation}
    W_{\mathrm{pre}}
    = U_{\mathrm{pre}}\Sigma_{\mathrm{pre}}V_{\mathrm{pre}}^\top,
    \qquad
    W_{\mathrm{ft}}
    = U_{\mathrm{ft}}\Sigma_{\mathrm{ft}}V_{\mathrm{ft}}^\top ,
\end{equation}
with singular vectors ordered by decreasing singular value. The first
$k$ columns of $U$ and $V$ define the dominant left and right singular
subspaces. Their principal-angle cosines are given by
\begin{equation}
    \cos \theta_i^{(L)}
    =
    \sigma_i\!\left(
        U_{\mathrm{pre},k}^{\top}U_{\mathrm{ft},k}
    \right),
    \qquad
    \cos \theta_i^{(R)}
    =
    \sigma_i\!\left(
        V_{\mathrm{pre},k}^{\top}V_{\mathrm{ft},k}
    \right),
\end{equation}

where $\sigma_i(\cdot)$ denotes the $i$-th singular value (Appendix~\ref{app:ssi_comp_detail}). We refer to the average overlap of the left and right subspaces as the
Spectral Stability Index (SSI):
\begin{equation}
    \mathrm{SSI}_k
    =
    \frac{1}{2}
    \left[
    \frac{1}{k}\sum_{i=1}^{k}\cos^2\theta_i^{(L)}
    +
    \frac{1}{k}\sum_{i=1}^{k}\cos^2\theta_i^{(R)}
    \right].
    \label{eq:ssi}
\end{equation}
Each term in Eq.~(\ref{eq:ssi}) is the normalized subspace similarity
$\|U_{\mathrm{pre},k}^\top U_{\mathrm{ft},k}\|_F^2 / k$ of \citet{hu2022lora},
since $\|U_{\mathrm{pre},k}^\top U_{\mathrm{ft},k}\|_F^2 = \sum_{i=1}^{k}\cos^2\theta_i^{(L)}$
(and likewise for $V$). The right-hand term with $k=128$ is also the
eigenspace alignment score of \citet{liu2025lift}. We apply it to compare
each pretrained weight with its fine-tuned counterpart, and average over
the left and right singular subspaces, since fine-tuning can preserve one
while rotating the other.

$\mathrm{SSI}_k \in [0,1]$ measures the overlap between the dominant
pretrained and fine-tuned weight subspaces: values close to one indicate
strong preservation, while lower values indicate greater rotation away
from the pretrained subspace. We use $k=32$ to focus on the dominant
singular directions, which capture the strongest linear transformations
of each weight matrix. Increasing $k$ includes progressively weaker modes
and dilutes this focus. At $k=\min(m,n)$, at least one subspace overlap
becomes trivially equal to one because the corresponding subspaces span
the full ambient space. Results for $k\in\{64,128\}$ are given in Appendix~\ref{app:ssi_sensitivity}.
%
%
We fix $k$ across models, since the adaptation dimension is set by the task, not by model width \citep{aghajanyan2021intrinsic,hu2022lora}; a width-proportional $k$ largely preserves encoder ordering and does not change conclusions (Appendix~\ref{sec:prop_k}).

\subsection{Effective Rank of Adaptation (ERA)}
SSI measures how much of the pretrained weight subspace is preserved,
whereas ERA measures how many directions are involved in the weight update.
We denote the singular values of $\Delta W$ by
$\lambda_1 \geq \lambda_2 \geq \cdots$. The singular values are normalized as
$p_i = \lambda_i / \sum_j \lambda_j$, with entropy
$H = -\sum_i p_i \log p_i$. We use the entropy-based effective rank
$\exp(H)$ of \citet{roy2007effective}, normalized by $\min(m,n)$:
\begin{equation}
    \mathrm{ERA}
    =
    \frac{\exp(H)}{\min(m,n)}.
    \label{eq:era}
\end{equation}
The numerator provides a soft estimate of the number of active singular
directions in $\Delta W$, and dividing by the maximum possible rank makes
the score comparable across matrix sizes. Low ERA therefore indicates
that the update is concentrated in a small number of directions, while
high ERA indicates that the update is distributed across a larger
fraction of the available weight space. Related analyses measure the
rank of $\Delta W$ by the number of singular values needed to explain
90\% of its energy \citep{biderman2024lora} or by its numerical rank
\citep{liu2025lift}. Unlike these, the effective rank requires no
threshold.

\subsection{Joint Interpretation}
\label{sec:joint_interpretation}

The three diagnostics capture complementary aspects of adaptation: $\rho$ measures update magnitude, SSI measures preservation of dominant pretrained subspaces, and ERA measures the dimensional breadth of the update.
In practice, $\rho$ and SSI are correlated, but $\rho$ alone does not explain SSI: random updates of the same norm leave the dominant subspaces far more intact than fine-tuning does (Appendix~\ref{sec:null_baseline}).
We expect a reusable foundation model to provide a feature basis that can be reweighted and combined to solve new tasks with limited changes to the pretrained structure. High SSI with strong downstream performance is consistent with this expectation. Modest $\rho$ and low ERA further indicate small, concentrated adjustments, whereas low SSI and high ERA indicate broader reorganization. We interpret these diagnostics jointly with performance, since high SSI alone can also reflect insufficient adaptation.
All diagnostics are computed per weight matrix and aggregated over attention and MLP layers; details and the choice of $k$ are in Appendix~\ref{app:spectral_details}.

%% file: sections/04_experiment.tex
\section{Experiments}\label{sec:exp}
%
\input{tables/aurora_results}

We first examine the spectral diagnostics, then assess benchmark suitability and compare adaptation across EO and CV models.

Building on PANGAEA~\citep{marsocci2026pangaea}, our EO evaluation covers optical segmentation at different spatial resolutions with MADOS~\citep{kikaki2024mados} and FLAIR-2~\citep{garioud2023flair2}, temporal crop mapping with PASTIS~\citep{garnot2021pastis}, pixel-level time-series classification with TimeMatch~\citep{nyborg2022timematch}, and SAR segmentation with BraDD~\citep{karaman2023bradd}. We select CROMA~\citep{fuller2023croma}, DOFA~\citep{xiong2024dofa}, Prithvi-2~\citep{szwarcman2025prithvi}, TerraMind~\citep{jakubik2025terramind}, and SSL4EO-MAE~\citep{wang2023ssl4eo} to cover a broad range of EO pretraining approaches. On TimeMatch, we additionally evaluate Presto~\citep{tseng2023presto} and Tessera~\citep{feng2026tessera}, which have a native time-series path, and AnySat~\citep{astruc2025anysat}. The selection rationale and further encoders are detailed in Appendix~\ref{app:selection}. Additional experiments examine SSI rank and learning rate sensitivity, decoder capacity, convolutional architectures, and cross-modal transfer in Appendix~\ref{app:more_results}.

For the CV reference, we use three widely established vision backbones: CLIP~\citep{radford2021clip}, DINOv2~\citep{oquab2023dinov2}, and DINOv3~\citep{simeoni2025dinov3}. We evaluate them on standard classification benchmarks (Flowers-102, CIFAR-10, and Oxford-IIIT Pets)~\cite{nilsback2008flowers,krizhevsky2009cifar,parkhi2012pets} and on PASCAL VOC~\cite{everingham2010voc} for semantic segmentation.

We compare full fine-tuning (FT) with two parameter-efficient adaptation methods chosen for complementary reasons. SVF is particularly relevant to our analysis because it preserves the pretrained singular vectors and adapts only their associated singular values, directly testing whether a downstream task can be solved by reweighting existing pretrained directions. LoRA provides a widely adopted low-rank adaptation baseline that permits the model to introduce new update directions. Together, they contrast reuse of the pretrained spectral basis with more flexible low-rank adaptation. Spectral diagnostics characterize full fine-tuning, with SSI reported at $k=32$. 
We use standard learning rates and decay schedules and monitor validation performance to assess convergence. Full optimization details, as well as dataset splits, model configurations, and adaptation settings, are provided in Appendix~\ref{training_details}.
 The code will be made publicly available at a later date.

\subsection{Interpreting the Spectral Diagnostics}
\label{sec:interpreting_diagnostics}

We first examine whether SSI and ERA reflect meaningful differences in pretraining and downstream adaptation. We consider two complementary comparisons: changing the pretrained model while holding the downstream task fixed, and changing the downstream task while holding the pretrained model fixed.

\paragraph{SSI captures preservation of dominant pretrained subspaces.}
Moving from DINOv2 to DINOv3 increases SSI from $0.771$ to $0.951$ on MADOS, while test mIoU improves from $41.9$ to $64.9$, and from $0.400$ to $0.908$ on PASTIS, with mIoU increasing from $31.5$ to $33.6$ (Fig.~\ref{fig:ssi_era_interpretation}). These observations are consistent with stronger pretraining providing directions that require less modification during adaptation, although the comparison also includes differences in architecture and pretraining recipe. Aurora~\citep{bodnar2025aurora}, a large-scale weather foundation model, provides a complementary example: adapting its $0.25^\circ$ pretrained model to $0.1^\circ$ forecasting yields lower SSI than fine-tuning at $0.25^\circ$, in both attention ($0.992 \rightarrow 0.902$) and MLP layers ($0.997 \rightarrow 0.953$). Refer to Appendix~\ref{app:aurora} for additional Aurora experiments on other tasks.

\paragraph{ERA captures the dimensional breadth of adaptation.}
For DINOv2 and DINOv3, ERA increases from MADOS to PASTIS, from $0.386$ to $0.572$ and from $0.630$ to $0.763$, respectively (Fig.~\ref{fig:ssi_era_interpretation}). The temporal crop-mapping task therefore involves updates distributed across more singular directions than the single-frame segmentation task. Similarly, Aurora's higher-resolution adaptation increases ERA from $0.150$ to $0.528$ in attention and from $0.141$ to $0.762$ in MLP layers. Together, these examples illustrate the complementary roles of the diagnostics: SSI measures whether dominant pretrained subspaces persist, whereas ERA measures how broadly the update is distributed. High subspace preservation can coexist with a high-dimensional update, as observed for DINOv3 on PASTIS.

\subsection{Which Benchmarks Reveal the Value of Pretraining?}

Not every downstream benchmark is equally informative about the value of a foundation model. Geospatial foundation models are evaluated on a highly fragmented and inconsistently reported set of benchmarks, with little agreement on which tasks should be used for comparison \citep{corley2026no}. Moreover, relatively simple tasks may already be solved well by models trained from scratch, so high scores alone do not demonstrate that large-scale pretraining has produced useful transferable features. We therefore first assess whether each benchmark reveals a benefit from pretraining, and restrict our main analysis to those that do.

We compare pretrained and architecture-matched, randomly initialized models under full fine-tuning and SVF. We compute the normalized pretraining gain as $\mathrm{gain} = (s_{\mathrm{pre}} - s_{\mathrm{rand}})/(s_{\mathrm{pre}} - s_{\mathrm{chance}})$, where $s_{\mathrm{pre}}$ and $s_{\mathrm{rand}}$ are downstream scores (accuracy or mIoU, in percent) obtained with pretrained and random initialization, and $s_{\mathrm{chance}}$ is the corresponding chance-level score. This expresses the improvement over training from scratch as a fraction of the pretrained model's performance above chance.

Averaged across encoders, Fig.~\ref{fig:dataset} shows small gains on HLSBurnScars \citep{phillips2023hlsburnscars} and Sen1Floods11 \citep{bonafilia2020sen1floods11}, but clearer benefits on MADOS, FLAIR-2, BraDD, PASTIS, and TimeMatch under the evaluated training budgets. We therefore focus our main analysis on these five benchmarks. MADOS exhibits the largest gains under both adaptation methods, exceeding those on the more complex temporal task PASTIS. Its limited labeled training data may make pretrained features particularly valuable, suggesting that task complexity alone does not determine the benefit of pretraining.

\subsection{Main Results}
\label{sec:main_results}

\paragraph{EO models substantially reorganize pretrained subspaces.}
Across all EO benchmarks, full fine-tuning departs far from the pretrained representation (Tab.~\ref{tab:eo_optical}--\ref{tab:eo_timematch}). Averaged over datasets, SSI is only 0.543/0.548 for attention/MLP layers, whereas the vision models retain values close to one (Tab.~\ref{tab:cv_adaptation}), and ERA and relative weight change $\rho$ are several times higher. As in full fine-tuning of LLMs \citep{biderman2024lora}, MLP updates have a higher effective rank than attention updates, here in all 38 model--dataset pairs (Tab.~2--5). Although the benchmarks are not matched across domains, the gap is consistent: EO adaptation typically reorganizes the dominant pretrained subspaces instead of adjusting features within a stable basis.


\input{tables/eo_results}

\paragraph{Vision models illustrate the reuse regime.}
With strong pretraining, full fine-tuning largely preserves vision foundation models’ dominant subspaces. On VOC, DINOv3 has SSI of 0.998/0.999 (attention/MLP) and $\rho \leq 0.021$, indicating near-complete preservation. Its ERA values of 0.048/0.065 are approximately an order of magnitude below the corresponding EO means. LoRA and SVF reach 78.3 and 81.4 mIoU, respectively, versus 77.5 for full fine-tuning.
High SSI and small relative weight changes persist for DINOv3 even with $10\times$ more optimization steps and a larger initial learning rate (Appendix~\ref{app:training-budget}).

\paragraph{Stronger pretraining is associated with more concentrated updates.}
Comparing DINOv2 with its stronger successor DINOv3 illustrates the effect of improved pretraining. DINOv3 consistently exhibits a clearer reuse signature, with higher SSI and lower $\rho$ and ERA. On CIFAR-10, for instance, SSI rises from 0.986/0.987 to 0.998/0.999, $\rho$ drops from 0.057/0.057 to 0.019/0.017, and ERA drops from 0.248/0.371 to 0.048/0.055, so the update is concentrated in several times fewer directions. This suggests that stronger pretraining yields a richer and more reusable basis, within which a downstream task can be solved by adjusting a few existing directions rather than introducing new ones. This agrees with \citet{turkoglu2026sve}, who show that SVF benefits increasingly from stronger pretraining, often exceeding full fine-tuning, also in the language domain.

\input{tables/cv_results}

\subsection{Cross-Domain Results with the Same Tasks and Training Protocol}

\label{sec:cross_domain_result}
Our main comparison evaluates EO and vision models in their respective
pretraining domains. This is the natural setting for a foundation model,
since transfer degrades with the distance between pretraining and target
data~\citep{matsoukas2022makes}, as the Aurora results also show (Sec.~\ref{sec:interpreting_diagnostics}).
%

As a complementary test, we fine-tuned both groups on the same EO tasks with the same input modality (RGB-only), same linear decoder, and training protocol (PANGAEA). This setting should favor the EO encoders because the target imagery is closer to their pretraining data. Even so, DINOv3 preserved more of its dominant weight subspaces (SSI are 0.972/0.755) than any EO encoder and outperformed the best EO model by 9.6 mIoU points on MADOS and 7.0 on FLAIR-2 (Tab.~\ref{tab:cross_domain_rgb}). DINOv2 had the lowest subspace overlap but still performs comparably with some EO models.

\subsection{Does Weight Subspace Overlap Reflect Feature Space Similarity?}

\label{sec:cka_analysis}
\begin{figure}[t]
\centering
\begin{minipage}[b]{0.49\linewidth}
\centering
{\small
\setlength{\tabcolsep}{2.5pt}
\begin{tabular}{@{}lrrrr@{}}
\toprule
 & \multicolumn{2}{c}{MADOS} & \multicolumn{2}{c}{FLAIR-2} \\
\cmidruletrim{lr}{2-3}\cmidruletrim{l}{4-5}
Encoder & $\mathrm{SSI}_{32}$ & mIoU & $\mathrm{SSI}_{32}$ & mIoU \\
\midrule
DINOv2     & .613 & 40.9 & .121 & 31.8 \\
DINOv3     & \textbf{.972} & \textbf{60.2} & \textbf{.755} & \textbf{46.7} \\
\midrule
CROMA      & .792 & 50.1 & .397 & 33.2 \\
DOFA       & .815 & 46.3 & .483 & 39.7 \\
Prithvi-2  & .886 & 49.6 & .486 & 37.4 \\
TerraMind  & .898 & 50.6 & .540 & 38.9 \\
SSL4EO-MAE & .839 & 40.9 & .485 & 32.8 \\
\bottomrule
\end{tabular}\par}
\captionof{table}{Spectral diagnostics of EO and CV models under the same tasks and training protocol.}
\label{tab:cross_domain_rgb}
\end{minipage}\hfill
\begin{minipage}[b]{0.45\linewidth}
\centering
\includegraphics[width=\linewidth]{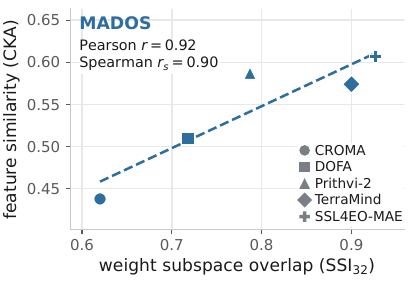}
\caption{CKA versus mean $\mathrm{SSI}_{32}$.}
\label{fig:ssi_cka}
\end{minipage}
\end{figure}

For the multispectral MADOS runs, we passed the same held-out validation
images through each EO encoder before and after full fine-tuning. At each
encoder block, we measured linear centered kernel alignment (CKA) between
the channel-standardized patch tokens of the two encoders~\citep{kornblith2019similarity}, and averaged the values over
blocks to obtain one score per encoder.
Across the five encoders, $\mathrm{SSI}_{32}$ correlated strongly with this mean CKA (Pearson $r=0.92$, Spearman $r_s=0.90$, Fig.~\ref{fig:ssi_cka}). This suggests that weight subspace overlap indicates how well an encoder preserves similarities between image patches in feature space during fine-tuning.

%% file: tables/aurora_results.tex

\begin{figure}[t]
\centering
\begin{minipage}[b]{0.44\linewidth}
  \centering
  \small
  \setlength{\tabcolsep}{6pt}
  \begin{tabular}{ll cc}
  \toprule
  & & \multicolumn{2}{c}{Resolution} \\
   &  & $0.25^\circ$ & $0.1^\circ$ \\
  \midrule
  \multirow{2}{*}{SSI}    & Attn & 0.992 & 0.902 \\
                          & MLP  & 0.997 & 0.953 \\
  \midrule
  \multirow{2}{*}{ERA}    & Attn & 0.150 & 0.528 \\
                          & MLP  & 0.141 & 0.762 \\
  \midrule
  \multirow{2}{*}{$\rho$} & Attn & 0.055 & 0.218 \\
                          & MLP  & 0.060 & 0.235 \\
\midrule
\multicolumn{2}{l}{Targets beating IFS HRES} & $>$91\% & 92\% \\
  \bottomrule
  \end{tabular}
  \captionof{table}{Aurora adaptation from the same $0.25^\circ$
  pretrained checkpoint. Higher-resolution forecasting exhibits
  lower SSI and higher ERA and relative weight change. Last row is taken from \citep{bodnar2025aurora}.}
  \label{tab:aurora}
\end{minipage}
\hfill
\begin{minipage}[b]{0.52\linewidth}
  \centering
  \includegraphics[width=\linewidth]{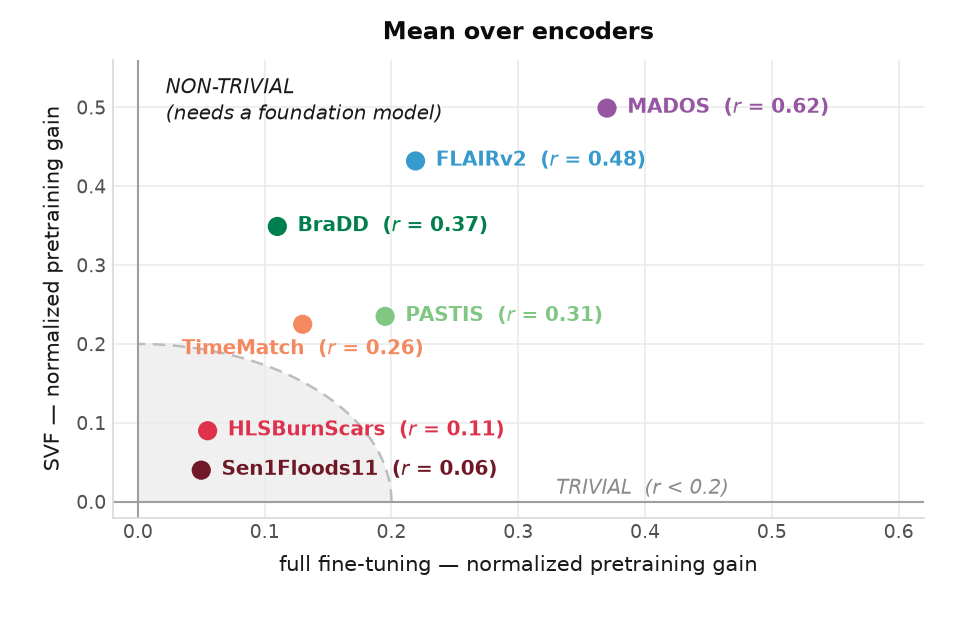}
  \caption{Normalized pretraining gains over architecture-matched random initialization under full fine-tuning and SVF, averaged across encoders.}
  \label{fig:dataset}
\end{minipage}
\end{figure}

%% file: tables/eo_results.tex
\begin{table*}[t]
\centering
\small
\setlength{\tabcolsep}{3pt}
\renewcommand{\arraystretch}{1.08}
\begin{tabular*}{\textwidth}{
@{\extracolsep{\fill}}ll cc cc cc ccc@{}
}
\toprule
& & \multicolumn{6}{c}{Full-FT spectral diagnostics}
& \multicolumn{3}{c}{mIoU (\%) $\uparrow$} \\
\cmidrule(lr){3-8}\cmidrule(l){9-11}
Dataset & Backbone
& \multicolumn{2}{c}{$\mathrm{SSI}_{32}$}
& \multicolumn{2}{c}{ERA}
& \multicolumn{2}{c}{$\rho$}
& Full & LoRA & SVF \\
\cmidrule(lr){3-4}\cmidrule(lr){5-6}\cmidrule(lr){7-8}
& & Attn & MLP & Attn & MLP & Attn & MLP & & & \\
\midrule

\multirow{6}{*}{MADOS}
& CROMA
& .545 & .695 & .466 & .542 & .452 & .462
& 63.6 & 63.9 & \textbf{64.4} \\
& DOFA
& .673 & .762 & .513 & .674 & .345 & .332
& 58.5 & 55.9 & \textbf{58.9} \\
& Prithvi-2
& .786 & .787 & .558 & .718 & .292 & .261
& \textbf{64.2} & 63.2 & 60.9 \\
& TerraMind
& .890 & .909 & .459 & .611 & .195 & .187
& \textbf{65.2} & 63.9 & 62.3 \\
& SSL4EO-MAE
& .930 & .923 & .324 & .437 & .136 & .141
& 46.1 & \textbf{46.9} & 41.9 \\
\cmidrule(l){2-11}
& \textit{Mean}
& .765 & .815 & .464 & .596 & .284 & .277
& \textbf{59.5} & 58.8 & 57.7 \\
\midrule

\multirow{6}{*}{PASTIS}
& CROMA
& .239 & .276 & .665 & .726 & .905 & .936
& \textbf{37.2} & 31.6 & 36.8 \\
& DOFA
& .684 & .764 & .584 & .723 & .372 & .370
& \textbf{39.9} & 39.0 & 34.3 \\
& Prithvi-2
& .460 & .378 & .745 & .874 & .602 & .604
& 33.7 & \textbf{37.0} & 33.9 \\
& TerraMind
& .919 & .895 & .612 & .759 & .202 & .240
& \textbf{43.2} & 41.6 & 38.9 \\
& SSL4EO-MAE
& .419 & .360 & .764 & .859 & .653 & .785
& \textbf{34.2} & 33.0 & 30.2 \\
\cmidrule(l){2-11}
& \textit{Mean}
& .544 & .535 & .674 & .788 & .547 & .587
& \textbf{37.6} & 36.4 & 34.8 \\
\midrule

\multirow{6}{*}{FLAIR-2}
& CROMA
& .348 & .446 & .525 & .616 & .606 & .694
& 33.2 & \textbf{34.0} & 31.8 \\
& DOFA
& .450 & .516 & .657 & .775 & .596 & .641
& 39.7 & \textbf{45.5} & 44.6 \\
& Prithvi-2
& .551 & .420 & .684 & .803 & .500 & .522
& 37.4 & \textbf{37.6} & 37.1 \\
& TerraMind
& .566 & .513 & .693 & .807 & .491 & .533
& 38.9 & \textbf{41.7} & 36.9 \\
& SSL4EO-MAE
& .522 & .448 & .699 & .783 & .518 & .646
& \textbf{32.8} & 28.7 & 21.6 \\
\cmidrule(l){2-11}
& \textit{Mean}
& .487 & .469 & .652 & .757 & .542 & .607
& 36.4 & \textbf{37.5} & 34.4 \\
\midrule

\multicolumn{2}{@{}l}{Overall mean}
& .599 & .606
& .597 & .714
& .458 & .490
& \textbf{44.5} & 44.2 & 42.3 \\
\bottomrule
\end{tabular*}

\caption{Full-FT spectral diagnostics and test mIoU on three EO
benchmarks. Attn and MLP denote layer groups, and $\rho$ denotes
relative weight change. Dataset means average the five backbones;
the overall mean weights all 15 model--dataset pairs equally.
Bold performance values identify the best adaptation method within
each backbone, and dataset mean.}
\label{tab:eo_optical}
\end{table*}


\begin{table*}[t]
\centering
\small
\setlength{\tabcolsep}{3pt}
\renewcommand{\arraystretch}{1.08}
\begin{tabular*}{\textwidth}{
@{\extracolsep{\fill}}ll cc cc cc ccc@{}
}
\toprule
& & \multicolumn{6}{c}{Full-FT spectral diagnostics}
& \multicolumn{3}{c}{mIoU (\%) $\uparrow$} \\
\cmidrule(lr){3-8}\cmidrule(l){9-11}
Dataset & Backbone
& \multicolumn{2}{c}{$\mathrm{SSI}_{32}$}
& \multicolumn{2}{c}{ERA}
& \multicolumn{2}{c}{$\rho$}
& Full & LoRA & SVF \\
\cmidrule(lr){3-4}\cmidrule(lr){5-6}\cmidrule(lr){7-8}
& & Attn & MLP & Attn & MLP & Attn & MLP & & & \\
\midrule

\multirow{5}{*}{BraDD}
& CROMA
& .287 & .357 & .672 & .743 & .914 & 1.090
& 54.1 & 55.8 & \textbf{57.4} \\
& DOFA
& .637 & .715 & .609 & .737 & .422 & .427
& \textbf{48.1} & 47.3 & 47.7 \\
& TerraMind
& .449 & .457 & .721 & .838 & .638 & .645
& 59.1 & 58.7 & \textbf{60.3} \\
& SSL4EO-MAE
& .745 & .694 & .685 & .796 & .349 & .400
& 52.3 & 51.8 & \textbf{53.7} \\
\cmidrule(l){2-11}
& Mean
& .530 & .556
& .672 & .779
& .581 & .641
& 53.4 & 53.4 & \textbf{54.8} \\
\bottomrule
\end{tabular*}

\caption{Full-FT spectral diagnostics and test mIoU on BraDD. Attn and MLP denote layer groups; $\rho$ is relative weight change. Mean row weights four backbones equally. \textbf{Bold} marks best per row.}
\label{tab:eo_bradd}
\end{table*}


\begin{table*}[t]
\centering
\small
\setlength{\tabcolsep}{3pt}
\renewcommand{\arraystretch}{1.08}
\begin{tabular*}{\textwidth}{
@{\extracolsep{\fill}}ll cc cc cc ccc@{}
}
\toprule
& & \multicolumn{6}{c}{Full-FT spectral diagnostics}
& \multicolumn{3}{c}{Accuracy (\%) $\uparrow$} \\
\cmidrule(lr){3-8}\cmidrule(l){9-11}
Dataset & Backbone
& \multicolumn{2}{c}{$\mathrm{SSI}_{32}$}
& \multicolumn{2}{c}{ERA}
& \multicolumn{2}{c}{$\rho$}
& Full & LoRA & SVF \\
\cmidrule(lr){3-4}\cmidrule(lr){5-6}\cmidrule(lr){7-8}
& & Attn & MLP & Attn & MLP & Attn & MLP & & & \\
\midrule

\multirow{8}{*}{TimeMatch}
& CROMA
& .116 & .074 & .589 & .628 & 1.330 & 1.445
& \textbf{85.6} & 85.4 & 73.7 \\
& DOFA
& .188 & .182 & .587 & .701 & .851 & .883
& \textbf{86.2} & 81.5 & 70.0 \\
& Prithvi-2
& .327 & .175 & .597 & .707 & .654 & .726
& \textbf{86.6} & 84.7 & 72.5 \\
& SSL4EO-MAE
& .437 & .310 & .696 & .712 & .639 & .834
& 84.3 & \textbf{85.5} & 73.0 \\
& AnySat
& .217 & .224 & .639 & .700 & .845 & .952
& \textbf{86.9} & 69.6 & 62.7 \\
& Presto
& .715 & .759 & .416 & .519 & 1.429 & 1.058
& \textbf{70.0} & 66.0 & 36.7 \\
& Tessera
& .716 & .836 & .760 & .840 & .424 & .406
& \textbf{87.9} & 84.9 & 72.6 \\
\cmidrule(l){2-11}
& Mean
& .388 &.366
& .612 & .687
& .882 & .901
& \textbf{83.9} & 79.7 & 65.9 \\
\bottomrule
\end{tabular*}

\caption{Full-FT spectral diagnostics and test accuracy on the canonical single-target TimeMatch experiment, with FR/31TCJ/2017 held out. Attn and MLP denote layer groups; $\rho$ is relative weight change. The mean row weights all seven backbones equally. Bold marks best performance per row.}
\label{tab:eo_timematch}
\end{table*}

%% file: tables/cv_results.tex
\begin{table*}[t]
\centering
\small
\setlength{\tabcolsep}{3pt}
\renewcommand{\arraystretch}{1.08}
\begin{tabular*}{\textwidth}{
@{\extracolsep{\fill}}ll cc cc cc ccc@{}
}
\toprule
& & \multicolumn{6}{c}{Full-FT spectral diagnostics}
& \multicolumn{3}{c}{Performance (\%) $\uparrow$} \\
\cmidrule(lr){3-8}\cmidrule(l){9-11}
Dataset & Backbone
& \multicolumn{2}{c}{$\mathrm{SSI}_{32}$}
& \multicolumn{2}{c}{ERA}
& \multicolumn{2}{c}{$\rho$}
& Full & LoRA & SVF \\
\cmidrule(lr){3-4}\cmidrule(lr){5-6}\cmidrule(lr){7-8}
& & Attn & MLP & Attn & MLP & Attn & MLP & & & \\
\midrule

\multirow{3}{*}{Flowers-102}
& CLIP
& .982 & .992 & .160 & .246 & .051 & .051
& 91.2 & 90.7 & \textbf{93.7} \\
& DINOv2
& .994 & .990 & .160 & .286 & .045 & .046
& 97.0 & \textbf{98.7} & 98.1 \\
& DINOv3
& .998 & .999 & .069 & .102 & .021 & .020
& \textbf{98.1} & 97.4 & 94.7 \\
\midrule

\multirow{3}{*}{CIFAR-10}
& CLIP
& .976 & .984 & .237 & .337 & .065 & .061
& \textbf{98.1} & 97.2 & 97.6 \\
& DINOv2
& .986 & .987 & .248 & .371 & .057 & .057
& 98.8 & 98.5 & \textbf{99.0} \\
& DINOv3
& .998 & .999 & .048 & .055 & .019 & .017
& 98.7 & 98.7 & \textbf{99.0} \\
\midrule

\multirow{3}{*}{Pets}
& CLIP
& .944 & .969 & .310 & .493 & .105 & .099
& 91.3 & \textbf{93.9} & 87.9 \\
& DINOv2
& .966 & .972 & .337 & .584 & .101 & .100
& 93.1 & \textbf{95.2} & \textbf{95.2} \\
& DINOv3
& .996 & .997 & .136 & .218 & .034 & .033
& 94.8 & 94.4 & \textbf{95.4} \\
\midrule


\multirow{3}{*}{VOC}
& CLIP
& .995 & .998 & .102 & .144 & .027 & .030
& 71.7 & 70.3 & \textbf{72.9} \\
& DINOv2
& .998 & .998 & .090 & .143 & .024 & .026
& \textbf{79.6} & 77.8 & 75.9 \\
& DINOv3
& .998 & .999 & .048 & .065 & .021 & .020
& 77.5 & 78.3 & \textbf{81.4} \\
\midrule


\multicolumn{2}{@{}l}{Overall mean}
& .986 & .990
& .162 & .254
& .048 & .047
& -- & -- & -- \\
\bottomrule
\end{tabular*}

\caption{Full-FT spectral diagnostics and downstream accuracy (mIoU for VOC). Attn and MLP denote layer groups; $\rho$ is relative weight change. Spectral means weight datasets and backbones equally. Bold marks the best performance per row.} 
\label{tab:cv_adaptation}
\end{table*}

%% file: sections/06_discussion.tex
\section{Discussion and Future Perspective}\label{sec:discussion}

\begin{figure*}
    \centering
    \includegraphics[width=\textwidth]{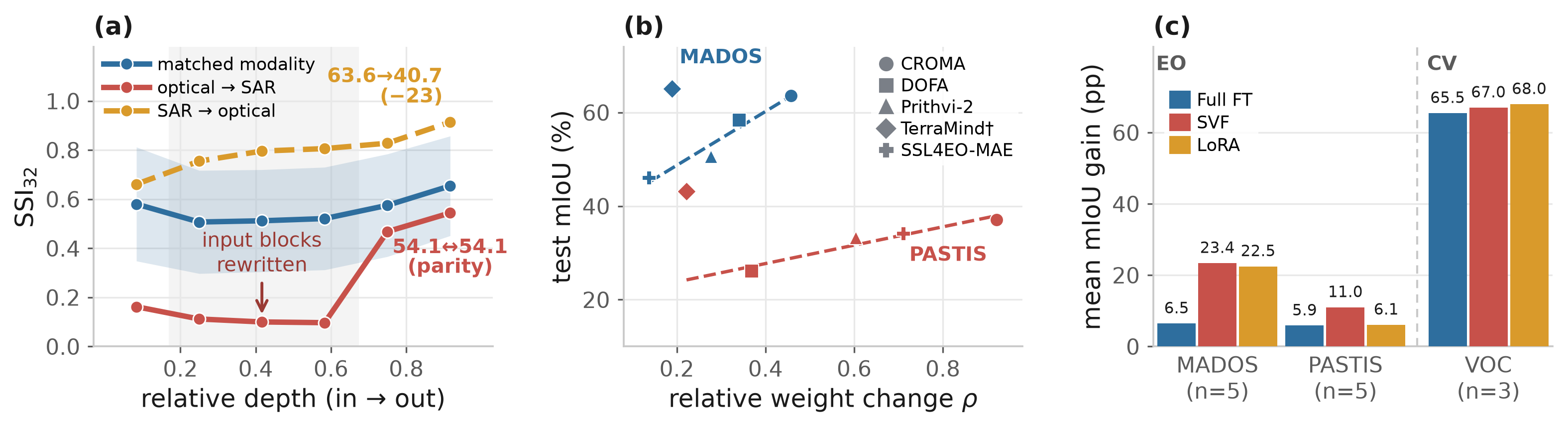}
    \caption{(a) Depth-resolved $\mathrm{SSI}_{32}$ under full-FT for matched-modality transfer (mean $\pm$1 std across encoders and benchmarks) and CROMA cross-modal transfer on BraDD. Annotations compare cross-modal and modality-matched mIoU. (b) Full-FT mIoU versus relative weight change $\rho$ on MADOS and PASTIS. Dashed lines are least-squares fits excluding TerraMind. (c) Test mIoU gains over architecture-matched random initialization, in pp, for full-FT, SVF, and LoRA: five EO backbones on MADOS/PASTIS and three CV backbones on VOC, under the evaluated training budgets.}
    \label{fig:adaptation_and_pretraining}
\end{figure*}

\paragraph{Reuse and relearning in EO and CV.}
The evaluated EO and CV models generally occupy different adaptation regimes. CV models largely preserve pretrained subspaces, whereas EO models undergo larger updates with higher effective ranks and greater subspace reorganization. EO pretraining thus often appears to provide a useful initialization for task-specific learning, with less readily reusable structure. The smaller pretraining gains on MADOS and PASTIS compared with VOC support this interpretation (Fig.~~\ref{fig:adaptation_and_pretraining}c), although benchmarks and training budgets differ.
\paragraph{A trade-off between performance and reuse.}
For most EO backbones, higher performance on MADOS and PASTIS is associated with larger relative weight changes (Fig.~\ref{fig:adaptation_and_pretraining}b). Together with their lower SSI, this suggests that these models often solve downstream tasks by substantially departing from their dominant pretrained subspaces. Such behavior may fall short of the foundation model premise of learning representations that can be readily reused across tasks. However, models achieving strong performance with limited changes show that this trade-off is not inevitable, highlighting the potential for more reusable EO pretraining.
\paragraph{Subspace preservation can coexist with broad adaptation.}
High SSI does not necessarily imply low-dimensional adaptation. On TimeMatch, Tessera \citep{feng2026tessera} achieves the highest accuracy while combining relatively high SSI with high ERA (Tab.~\ref{tab:eo_timematch}). This illustrates that successful adaptation can preserve much of the dominant pretrained subspaces while distributing updates across many directions. Subspace preservation and adaptation breadth therefore capture complementary aspects of transfer and should be interpreted jointly.

\paragraph{Parameter-efficient transfer.} The diagnostics may also indicate when a model can be adapted cheaply. On CV benchmarks, where SSI is close to $1$, LoRA or SVF mostly matches or exceeds full FT. On MADOS, the best-preserved EO benchmark, LoRA and SVF come within 0.7 and 1.8 points of full fine-tuning, with SVF updating under 0.1\% of the parameters (Tab.~\ref{tab:peft_budget}); on TimeMatch, where subspaces are largely rewritten, they fall 4.2 and 18.0 points behind. This holds on average but not per backbone; a per-model link needs method-specific tuning, left for future work.

\paragraph{Cross-modal transfer and adaptation depth.}
Subspace changes generally concentrate in early and middle layers, with larger preservation toward the output (Fig.~\ref{fig:adaptation_and_pretraining}a). CROMA's cross-modal experiments illustrate why performance and reuse must be distinguished (Appendix~\ref{sec:abl-xmodal}). On BraDD, the optical encoder matches the SAR encoder at 54.1 mIoU while substantially reorganizing its early and middle layers. Conversely, the SAR encoder preserves more pretrained structure on MADOS but reaches only 40.7 mIoU, versus 63.6 for the optical encoder. Strong performance can thus coexist with extensive rewriting, and high preservation with poor transfer: neither accuracy nor SSI alone establishes successful reuse.


\paragraph{Limitation.}
Our main analyses use fixed fine-tuning configurations, complemented by targeted learning-rate sweeps for one CV and two EO encoders (Appendix~\ref{app:lr_sensitivity}). These sweeps support our main conclusions but also show that a shared learning rate can under-train individual encoders. Extending them to all configurations was beyond our computational budget, as each configuration would require $10\times$ the fine-tuning compute. Future work should broaden these sweeps and explore aggregate measures, such as normalized area-under-the-curve scores over a shared log-scaled learning-rate range, to assess adaptation more broadly than a single configuration.

\paragraph{Future work.}
Our diagnostics could guide the design of pretraining objectives that produce more broadly reusable representations, supporting strong downstream performance with less structural reorganization. Controlled studies varying pretraining data size and diversity, architecture, model capacity, and compute budget could further disentangle their contributions to reuse and investigate whether representation reuse follows predictable scaling relationships.

%% file: sections/02_RW.tex
\section{Related Work}\label{sec:RW}

\paragraph{Linear subspaces as feature bases.}
Recent work suggests that knowledge in neural networks is organized along meaningful linear subspaces. In transformer weight matrices, singular vectors have been associated with semantically meaningful directions in representation space \citep{elhage2022superposition,millidge2022svd}, while analyses of their spectra suggest that information is distributed across singular modes \citep{staats2025small}. Related work on model editing further shows that targeted low-rank changes to weight matrices can modify specific learned associations \citep{meng2022rome}. Fine-tuning itself has also been shown to operate in a low-dimensional intrinsic subspace
\citep{aghajanyan2021intrinsic}. Methods such as Singular Value Fine-tuning build on this perspective by preserving pretrained singular directions while adapting their associated singular values \citep{sun2022svf,turkoglu2026sve}. Related analyses of transfer learning have examined how pretrained models change during adaptation in feature and parameter space, including the spectral structure of individual modules \citep{neyshabur2020being}. Together, these findings motivate treating the dominant singular subspace of a pretrained model as a
meaningful feature basis whose preservation can be studied during downstream adaptation.

\paragraph{Spectral structure of model adaptation.}
Several parameter-efficient methods exploit low-rank or spectral structure during adaptation. LoRA constrains weight updates to be low rank \citep{hu2022lora}, while Spectral Adapter and related methods parameterize adaptation directly in the singular basis \citep{zhang2024spectraladapter,zhao2025sst}. 
Recent work has also used spectral structure to analyze fine-tuning itself. \citet{si2025weight} study changes in singular values and singular-vector alignment between pretrained and fine-tuned weights. Similarly, \citet{shuttleworth2025lora} show that LoRA and full fine-tuning can reach similar downstream performance while inducing substantially different singular-vector structure, including LoRA-specific ``intruder'' directions. \citet{biderman2024lora} show that full fine-tuning of LLMs produces high-rank updates, with higher ranks in MLP than in attention modules. Relatedly, Diffract studies the spectral evolution of LLMs during continued pretraining and finds that substantial adaptation can occur through changes
in singular directions despite comparatively stable singular-value spectra \citep{borodin2026diffract}. Our objective is complementary: rather than constraining adaptation or tracking individual singular directions, we use spectral structure to \emph{audit representation reuse}. SSI measures preservation of the dominant pretrained left and right singular subspaces through principal angles, while ERA characterizes the effective dimensionality of the update $\Delta W$.

\paragraph{Feature reuse and fine-tuning dynamics.}
Under domain shift, \citet{raghu2019transfusion} found that transfer from ImageNet to medical imaging brings little benefit, with feature reuse concentrated in the lowest layers and part of the benefit stemming from the scale of the pretrained weights. Finetuning can also distort pretrained features when a new head is trained jointly with the encoder \citep{kumar2022finetuning}, and finetuning BERT mainly changes top layers \citep{merchant2020happens}.

\paragraph{Foundation models for EO and Earth system.}
Foundation models aim to learn representations that can be reused across downstream tasks with limited adaptation \citep{bommasani2021foundation}. This paradigm, established in language and general vision \citep{brown2020language,oquab2023dinov2}, has increasingly been adopted in Earth observation (EO). SatMAE targets temporal and multispectral imagery \citep{cong2022satmae}, CROMA jointly learns from optical and SAR observations \citep{fuller2023croma}, and Prithvi-EO-2.0 targets transfer across multitemporal EO applications \citep{szwarcman2025prithvi}. At the same time, benchmark studies show that geospatial foundation models do not consistently outperform conventional or task-specific alternatives \citep{marsocci2026pangaea,corley2026no}. \citet{sosa2024effective} similarly find that MAE pretraining helps most when the downstream task resembles the pretraining task, and that tuned training from scratch can match it for segmentation and classification. Related developments in weather and Earth-system modelling range from data-driven forecasters such as GraphCast \citep{lam2023graphcast} to broadly adaptable models such as Aurora \citep{bodnar2025aurora}. These developments motivate evaluating not only downstream accuracy, but also whether adaptation actually reuses the pretrained representation.

%% file: sections/05_conclusion.tex
\section{Conclusion}\label{sec:conclusion}

  We used established spectral measures to examine representation reuse through subspace preservation, update magnitude, and adaptation breadth. Across the evaluated benchmarks, EO models generally undergo larger, higher-rank updates and greater subspace reorganization than natural-image models, showing that strong downstream performance can coexist with extensive rewriting of pretrained structure. 
  Even when finetuned on the same EO tasks with the same protocol, DINOv3 preserves
more of its pretrained weight subspaces than every EO encoder and reaches
higher accuracy.
  Our findings motivate evaluating foundation models through downstream performance, pretraining benefits, and structural preservation together, since neither accuracy nor preservation alone establishes successful pretraining representation reuse. This perspective can guide EO pretraining toward representations that support strong transfer with less relearning.

%% file: appendix.tex

\section{Additional Details on Spectral Diagnostics}
\label{app:spectral_details}

\subsection{Matrix Extraction and Aggregation}
\label{app:matrix_extraction}

We compute the spectral diagnostics independently for each analyzed
weight matrix in the encoder. Attention query, key, value, and output
projections are grouped as \textsc{Attn}, while feed-forward layers are
grouped as \textsc{MLP}. Task-specific heads and normalization parameters
are excluded because they do not have a directly comparable pretrained
counterpart. For convolutional encoders, convolutional kernels of shape
$[C_{\mathrm{out}}, C_{\mathrm{in}}, k_h, k_w]$ are reshaped to
$[C_{\mathrm{out}}, C_{\mathrm{in}}k_hk_w]$ before computing the
diagnostics. Diagnostics are computed per matrix and subsequently
aggregated within the corresponding layer group.



\subsection{Computing SSI Principal Angles}\label{app:ssi_comp_detail}
Let
$U_{\mathrm{pre},k},U_{\mathrm{ft},k}\in\mathbb{R}^{m\times k}$
contain the top-$k$ left singular vectors of $W_{\mathrm{pre}}$ and
$W_{\mathrm{ft}}$, and let
$V_{\mathrm{pre},k},V_{\mathrm{ft},k}\in\mathbb{R}^{n\times k}$
contain the corresponding right singular vectors. Their columns are
orthonormal and therefore span $k$-dimensional subspaces. The principal
angles $0\leq\theta_1\leq\cdots\leq\theta_k\leq\pi/2$ between two such
subspaces are obtained from the singular values of the cross-Gram matrix
between their orthonormal bases \citep{bjorck1973numerical}:
\begin{equation}
    \cos\theta_i^{(L)}
    =
    \sigma_i\!\left(
        U_{\mathrm{pre},k}^{\top}U_{\mathrm{ft},k}
    \right),
    \qquad
    \cos\theta_i^{(R)}
    =
    \sigma_i\!\left(
        V_{\mathrm{pre},k}^{\top}V_{\mathrm{ft},k}
    \right).
\end{equation}
Each $\cos\theta_i\in[0,1]$ measures the overlap of a principal
direction between the two subspaces. The principal angles depend only
on the subspaces and are invariant to the particular orthonormal bases
used to represent them.

\subsection{Choice of SSI Subspace Rank $k$}
\label{app:ssi_rank}




\paragraph{Why truncate the singular subspaces?}
SSI measures whether fine-tuning preserves the dominant pretrained
subspaces. Truncation is essential because a complete basis spans the
same ambient space regardless of how its vectors rotate. At
$k=\min(m,n)$, at least one subspace overlap is therefore identically
one; for square matrices, $\mathrm{SSI}_k=1$ regardless of the update.
Restricting the comparison to the leading $k$ singular directions
avoids this degeneracy and focuses the diagnostic on the strongest
linear transformations represented by each matrix. Related spectral
analyses also use truncation: Diffract~\citep{borodin2026diffract},
for example, assesses adaptation redundancy by retaining only the
leading singular components of the weight update $\Delta W$ and
evaluating the resulting model.

\paragraph{Choice of $k$.}
We use $k=32$ across architectures to compare the same number of
dominant directions while remaining below full rank for every
analyzed matrix. Larger $k$ progressively includes weaker singular
modes and broadens the comparison beyond the dominant subspaces.
We report $k\in\{64,128\}$ as sensitivity checks.

\section{Additional Results and Ablations}
\label{app:more_results}

\subsection{Aurora Results}
\label{app:aurora}

Aurora provides a useful within-model comparison because the same pretrained backbone is adapted to several operational forecasting settings \citep{bodnar2025aurora}. These include weather forecasting at the pretraining resolution of $0.25^\circ$, higher-resolution weather forecasting at $0.1^\circ$, ocean-wave prediction at $0.25^\circ$, and air-pollution forecasting at $0.4^\circ$. The resulting spectral changes are consistent with the increasing departure from the original forecasting setting. Adaptation at $0.25^\circ$ leaves the pretrained subspaces almost unchanged, with SSI above $0.99$ and small ERA and relative weight change. Moving to $0.1^\circ$ weather forecasting produces substantially lower SSI and higher ERA and $\rho$, indicating that higher spatial resolution requires a broader reorganization of the pretrained weights. Wave and air-pollution forecasting similarly induce larger, higher-rank updates, with the strongest changes observed for air pollution. These results provide an additional within-architecture sanity check that SSI and ERA respond meaningfully to changes in downstream task and resolution.

\begin{table}[h]
\tiny
\centering
\begin{tabular}{l l c l cc cc cc}
\hline
 &  &  &  & \multicolumn{2}{c}{\textbf{SSI}} & \multicolumn{2}{c}{\textbf{ERA}} & \multicolumn{2}{c}{\textbf{Rel. Change}} \\
\cmidrule(lr){5-6} \cmidrule(lr){7-8} \cmidrule(lr){9-10}
\textbf{Downstream Model} & \textbf{Base Model} & \textbf{Res.} & \textbf{Task} 
& \textbf{Attn} & \textbf{MLP} 
& \textbf{Attn} & \textbf{MLP} 
& \textbf{Attn} & \textbf{MLP} \\
\hline
Aurora 0.25° Fine-Tuned & Aurora 0.25° Pretrained & 0.25° & HRES T0 Forecast & 0.992 & 0.997 & 0.150 & 0.141 & 0.055 & 0.060 \\
Aurora 0.1° Fine-Tuned & Aurora 0.25° Pretrained & 0.1° & High-Res Forecast & 0.902 & 0.953 & 0.528 & 0.762 & 0.218 & 0.235 \\
Aurora 0.25° Wave & Aurora 0.25° Pretrained & 0.25° & Wave Prediction & 0.918 & 0.945 & 0.563 & 0.779 & 0.220 & 0.277 \\
Aurora 0.4° Air Pollution & Aurora 0.25° 12h Pretrained & 0.4° & Air Pollution & 0.865 & 0.917 & 0.683 & 0.864 & 0.319 & 0.356 \\
\hline
\end{tabular}
\caption{Spectral diagnostics for Aurora (1.3B) across downstream Earth-system tasks.}
\end{table}

\subsection{Sensitivity to the SSI Subspace Rank}
\label{app:ssi_sensitivity}

\begin{figure}[t]
    \centering
    \includegraphics[width=\linewidth]{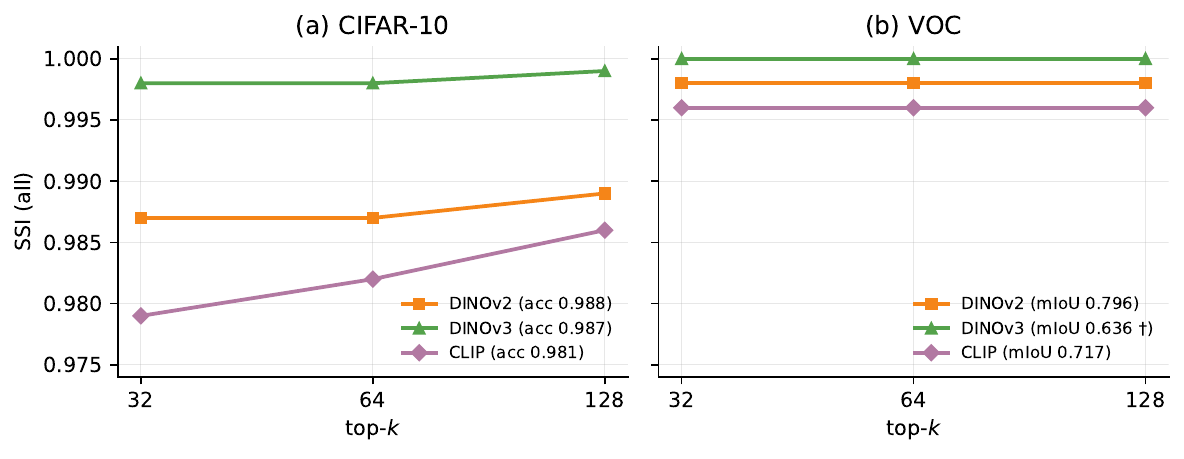}
\caption{Sensitivity of SSI to the subspace rank $k$ on (a) CIFAR-10 and (b) VOC. SSI remains largely stable across
the evaluated ranks, showing that high subspace preservation is robust to the choice of $k$ for the runs shown. Legends report accuracy and mIoU, respectively.}

    \label{fig:ssi_k_cv}
\end{figure}

\begin{figure}[t]
    \centering
    \includegraphics[width=\linewidth]{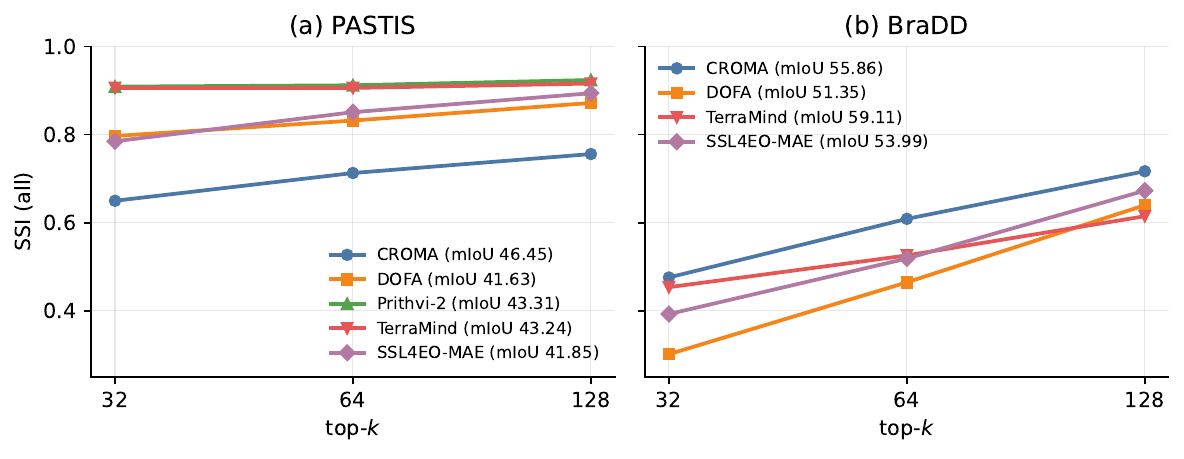}
\caption{Sensitivity of SSI to rank $k$ on (a) PASTIS and (b) BraDD. Legends report mIoU. Each curve varies $k$ with trained weights fixed.}

    \label{fig:ssi_k_eo}
\end{figure}

We vary $k\in\{32,64,128\}$ to assess the stability of SSI and
its interpretation. For CV backbones (Fig.~\ref{fig:ssi_k_cv}),
SSI remains close to one across ranks, with little change in
absolute values and an unchanged model ordering. Among EO encoders
(Fig.~\ref{fig:ssi_k_eo}), SSI is comparatively stable on PASTIS,
particularly for models with high preservation. On BraDD, scores
increase more noticeably, but all models follow similar upward
trends. Although individual EO rankings can vary, the broad
preservation patterns remain consistent. The qualitative conclusions
are therefore robust across the evaluated choices of $k$.

Increasing $k$ includes progressively weaker singular modes and
expands the compared subspaces, which can raise their overlap and
compress differences between models. We therefore choose $k=32$
as a practical trade-off: it summarizes dominant weight structure
across multiple singular directions while retaining sensitivity
to differences in subspace preservation. The larger-rank
evaluations show that the overall interpretation extends beyond
this particular choice.

\subsection{Width-Proportional Rank}
\label{sec:prop_k}
A fixed $k$ covers a different fraction of the spectrum for models of
different width $d$. We therefore recompute SSI with a width-proportional
rank, $k_{\mathrm{prop}} = \mathrm{round}(32\,d/1024)$, so that every
model keeps the same fraction of its width ($k_{\mathrm{prop}}=32$ for
ViT-L, $24$ for ViT-B, $12$ for ViT-S). Table~\ref{tab:prop_k} reports
the resulting change in SSI on MADOS. The differences are small ($-0.027$ on average), and the ordering of the encoders is largely preserved on every benchmark (rank correlation between the two versions is at least 0.90).
For models narrower than ViT-L, the proportional rule uses fewer than 32
directions, which lowers their SSI. The fixed $k=32$ therefore gives
narrower models slightly higher SSI than the proportional rule would. The only large change occurs for Presto on
TimeMatch ($-0.52$): its width is only $128$, so the proportional rule
keeps just $4$ directions.

\begin{table}[h]
\centering
\small
\begin{tabular}{@{}lrrr@{}}
\toprule
Encoder & Width $d$ & $k_{\mathrm{prop}}$ & $\Delta\,\mathrm{SSI}$ \\
\midrule
TerraMind  & 768  & 24 & $-0.004$ \\
Prithvi-2  & 768  & 24 & $-0.008$ \\
SSL4EO-MAE & 384  & 12 & $-0.074$ \\
DOFA       & 768  & 24 & $-0.050$ \\
CROMA      & 1024 & 32 & $0.000$  \\
\midrule
Mean       &      &    & $-0.027$ \\
\bottomrule
\end{tabular}
\caption{Change in SSI when moving from $k=32$ to the width-proportional
rank $k_{\mathrm{prop}}$ (MADOS).}
\label{tab:prop_k}
\end{table}

\subsection{Learning Rate Sensitivity and Under-Training}
\label{app:lr_sensitivity}

\begin{figure}[t]
    \centering
    \begin{minipage}{0.49\linewidth}
        \centering
        \includegraphics[width=\linewidth]{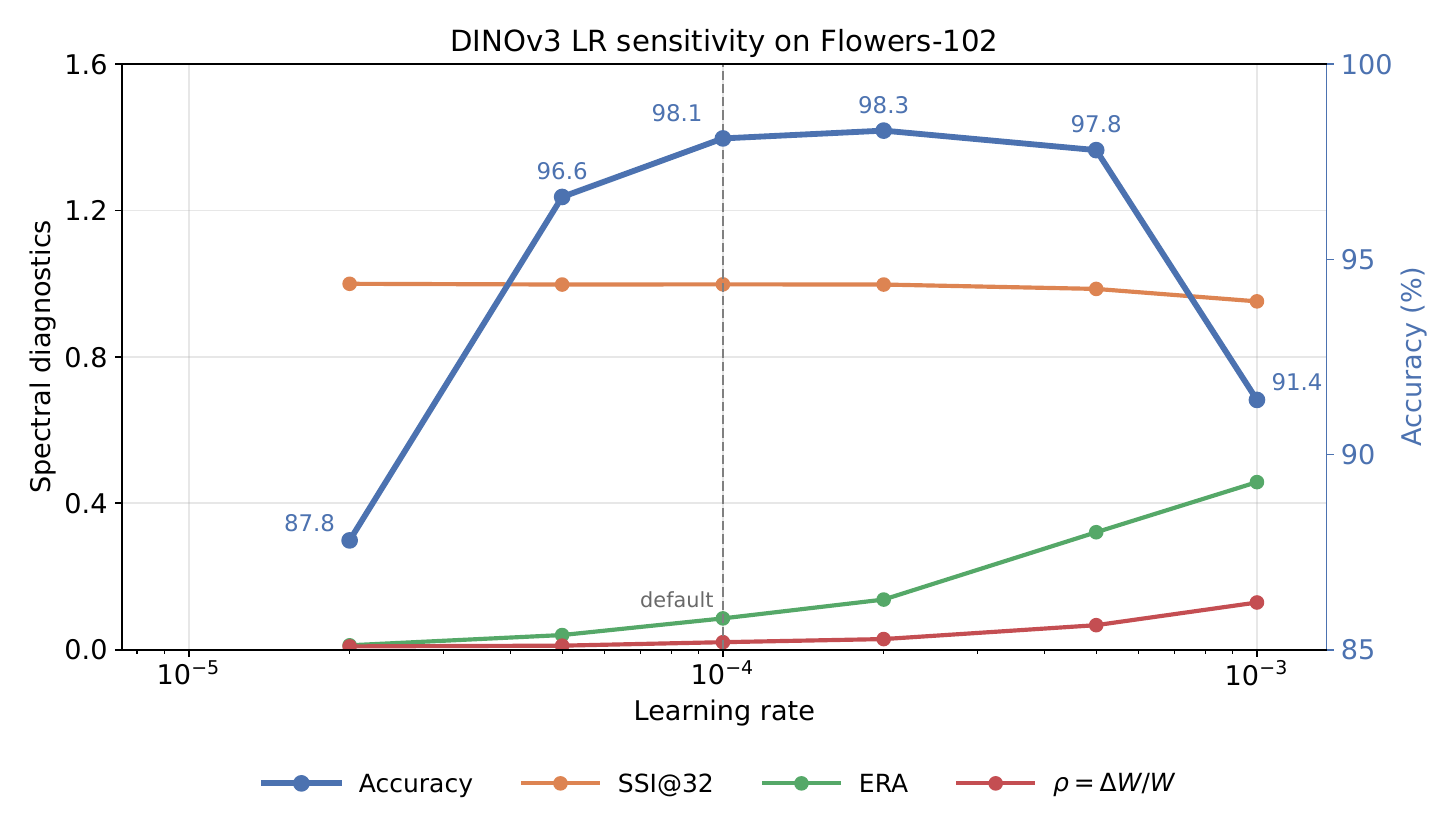}
        \textbf{(a)} Flowers-102
    \end{minipage}
    \hfill
    \begin{minipage}{0.49\linewidth}
        \centering
        \includegraphics[width=\linewidth]{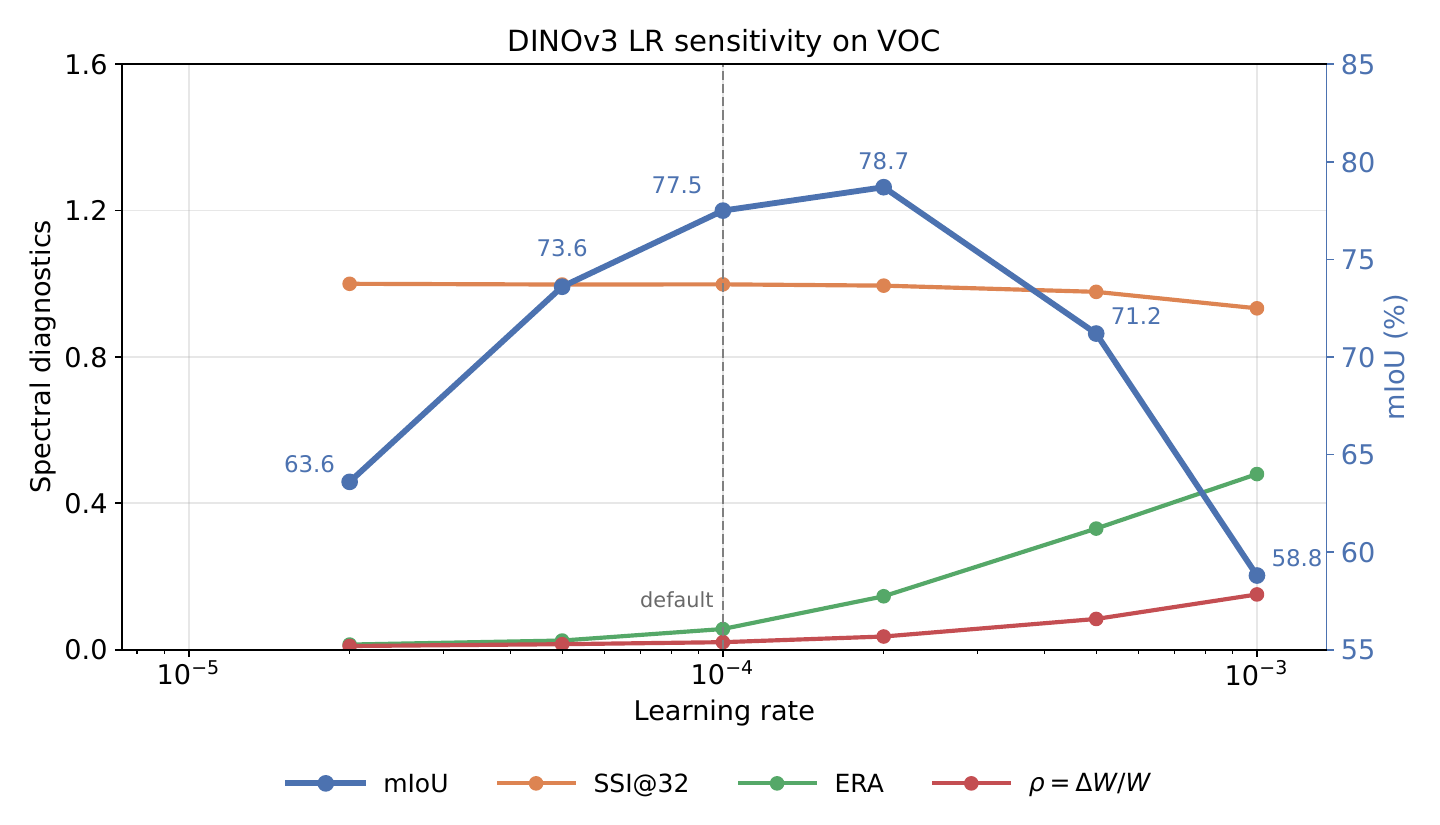}
        \textbf{(b)} VOC
    \end{minipage}

\caption{Learning-rate sensitivity of DINOv3 under full fine-tuning on (a) Flowers-102 and (b) VOC. Spectral diagnostics use the left axis and test performance the right axis; the dashed line marks the default learning rate. At low learning rates, the encoder changes very little and SSI remains close to one despite suboptimal performance. Performance peaks at $10^{-4}$ to $2\times10^{-4}$, where SSI$_{32}$ remains above 0.99. Larger learning rates degrade performance and broaden the update (higher ERA), but relative weight change stays small ($\rho \leq 0.15$) and SSI$_{32}$ stays above 0.93. Overall, subspace preservation is stable across the sweep, unlike for the EO encoders (Fig.~\ref{fig:lr_sensitivity_EO}).}

    \label{fig:lr_sensitivity}
\end{figure}

\begin{figure}[t]
    \centering
    \begin{minipage}{0.49\linewidth}
        \centering
        \includegraphics[width=\linewidth]{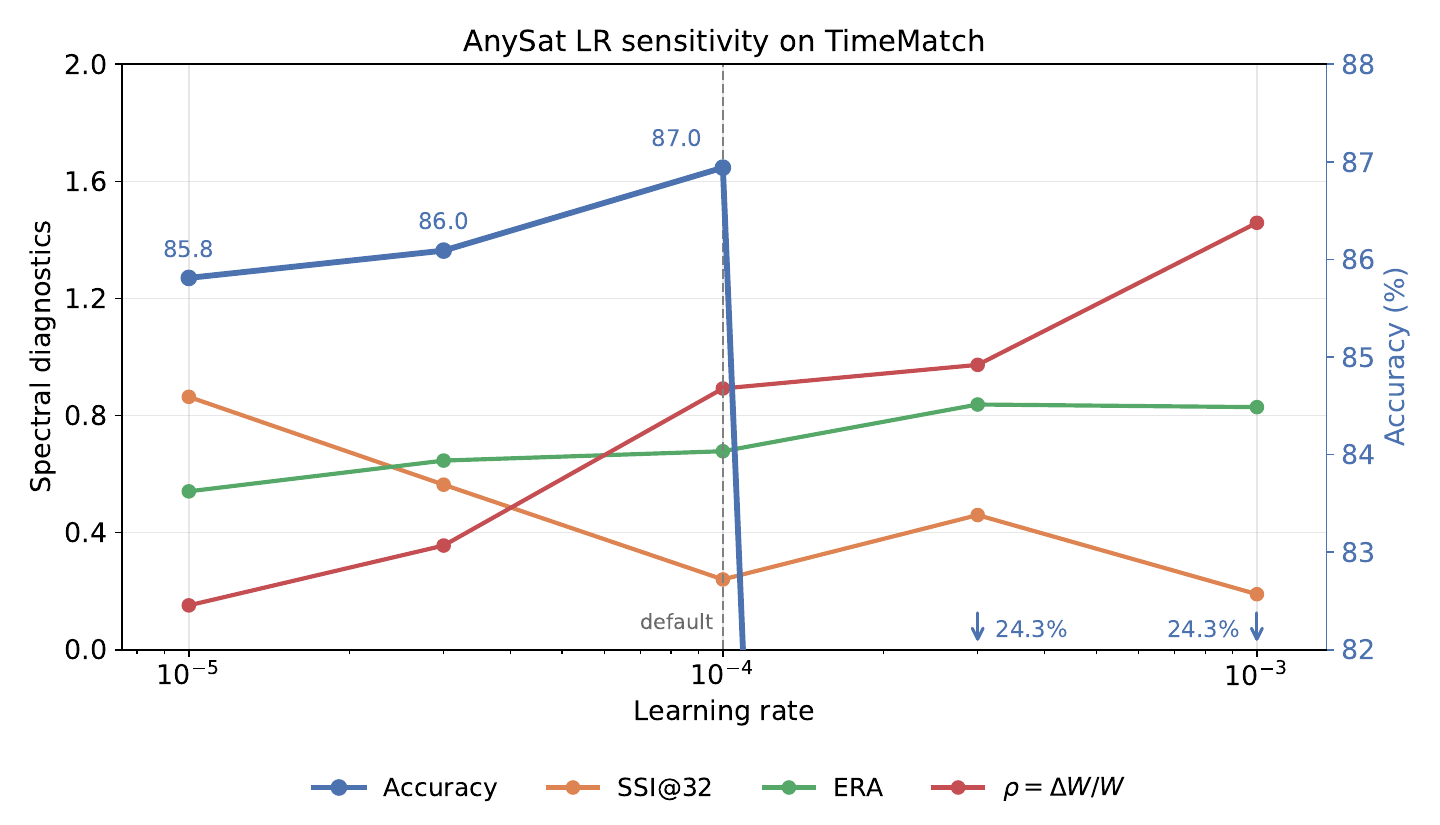}
        \textbf{(a)} AnySat
    \end{minipage}
    \hfill
    \begin{minipage}{0.49\linewidth}
        \centering
        \includegraphics[width=\linewidth]{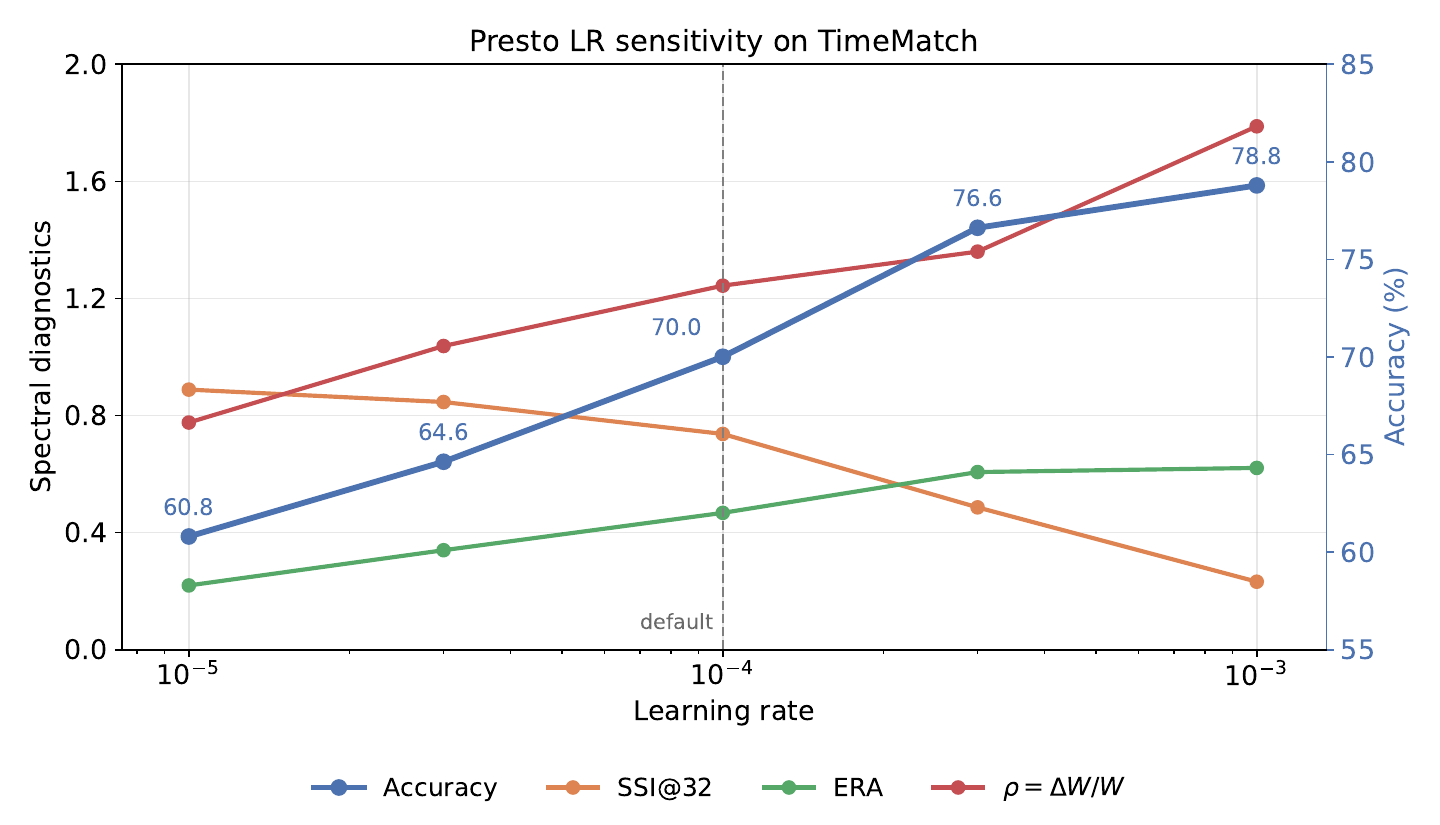}
        \textbf{(b)} Presto
    \end{minipage}

   \caption{Learning rate sensitivity of (a) AnySat and (b) Presto under full fine-tuning on TimeMatch. Spectral diagnostics use the left axis and test accuracy the right axis; the dashed line marks the default learning rate used in Tab.~\ref{tab:eo_timematch}. Unlike DINOv3 (Fig.~\ref{fig:lr_sensitivity}), neither encoder reaches its best accuracy while preserving its dominant subspaces. AnySat peaks at the default with SSI$_{32}$ of 0.24; lower learning rates raise SSI at a cost in accuracy, and learning rates of $3\times10^{-4}$ and above diverge (24.3\%, below the plotted range). Presto improves monotonically with the learning rate while SSI decreases and $\rho$ increases, so its best accuracy coincides with the lowest SSI in the sweep. Overall, the EO diagnostics are far less stable across learning rates than their CV counterparts.}
    \label{fig:lr_sensitivity_EO}
\end{figure}

A high SSI can reflect either useful preservation or insufficient
adaptation. We examine this distinction by sweeping the full
fine-tuning learning rate for DINOv3 on Flowers-102 and VOC
(Fig.~\ref{fig:lr_sensitivity}) and for AnySat \citep{astruc2025anysat} and Presto \citep{tseng2023presto} on
TimeMatch (Fig.~\ref{fig:lr_sensitivity_EO}). For DINOv3, the spectral
diagnostics remain largely stable across the sweep: over almost two
orders of magnitude in learning rate, SSI stays above 0.93 and $\rho$
at most 0.15, while ERA remains low up to the performance optimum.
Very low learning rates yield high SSI and small weight changes
despite suboptimal performance. Increasing the learning rate improves
performance until it reaches a plateau, over which SSI remains high
and largely stable. Higher learning rates broaden the update, with
higher ERA and poorer performance.

The EO encoders behave differently. Their spectral diagnostics shift
substantially across the sweep, with SSI falling from about 0.9 at
the lowest learning rate to below 0.25 within one to two orders of
magnitude. Presto improves monotonically with the learning rate
while SSI decreases and $\rho$ increases, so its best accuracy
coincides with the lowest SSI in the sweep and the default learning
rate under-trains it. AnySat reaches its best accuracy at the default
with low SSI. Lower learning rates raise SSI at a small cost in
accuracy, whereas higher learning rates cause training to diverge.
Neither encoder thus combines its best performance with preservation
of its dominant subspaces, and the gap between CV and EO adaptation
in Sec.~\ref{sec:main_results} persists at each encoder's best
learning rate.

These results motivate interpreting SSI, ERA, and $\rho$
jointly with downstream performance. In particular, stable SSI
across the performance plateau shows that substantial subspace
preservation can accompany successful adaptation. High SSI with
poor performance, however, may reflect insufficient learning, as
observed for Presto at low learning rates. Diagnostics at a single
learning rate therefore do not capture the full adaptation behavior
of a model. Nevertheless, the sweeps support the conclusions drawn
from the fixed configurations in our main analyses: the CV reference
preserves its dominant subspaces across a wide range of learning
rates, whereas the EO encoders reach their best performance only
with substantial rewriting. We discuss this limitation in
Section~\ref{sec:discussion}.

\subsection{Sensitivity to the Fine-Tuning Budget}
\label{app:training-budget}

To assess whether subspace preservation persists under more extensive
adaptation, we fine-tune DINOv3 on Flowers-102 and PASCAL VOC for
100 epochs, using $10\times$ more optimization steps than the default
10-epoch budget. We also double the initial learning rate from
$10^{-4}$ to $2\times10^{-4}$ to further test this behavior under
larger optimization steps.

\begin{figure}[t]
    \centering
    \includegraphics[width=\linewidth]{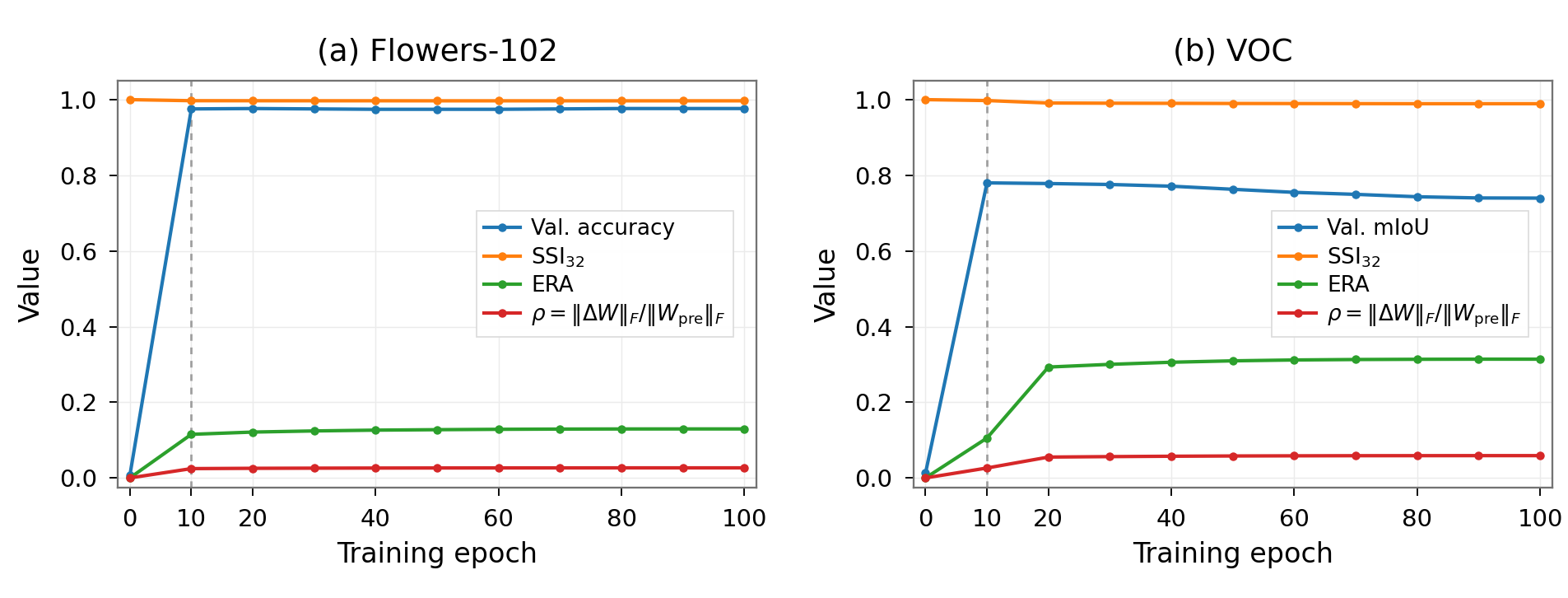}
    \caption{Extended full fine-tuning of DINOv3 on Flowers-102 and
    PASCAL VOC, using $10\times$ more optimization steps and twice the
    default initial learning rate. Curves show validation accuracy or
    mIoU as fractions, alongside SSI, ERA, and relative weight change
    $\rho$, aggregated over the analyzed encoder matrices. The dashed
    line marks epoch 10 within each extended run. SSI remains high
    and relative weight change remains small, while ERA increases
    more noticeably on VOC but still stays relatively low.}
    \label{fig:training-budget}
\end{figure}

Figure~\ref{fig:training-budget} shows the evolution of validation
performance and spectral diagnostics throughout these extended runs.
Between epochs 10 and 100, Flowers-102 accuracy remains similar
(97.55\% to 97.65\%), whereas VOC mIoU declines from 78.00\% to 73.97\%.
SSI remains high on both datasets, ending at 0.9969 and 0.9893,
respectively. Relative weight change increases only modestly in
absolute terms and remains small ($\rho=0.0266$ and $0.0589$).
ERA increases more noticeably on VOC, from 0.1054 to 0.3139, but
remains below the EO benchmark averages reported in our main analysis.
These results suggest that high subspace preservation and small
relative updates persist in these two settings despite substantially
longer training and a larger initial learning rate.

\subsection{Effect of Decoder Capacity}
\label{sec:abl-decoder}

\begin{figure}[t]
\centering
\includegraphics[width=0.62\textwidth]{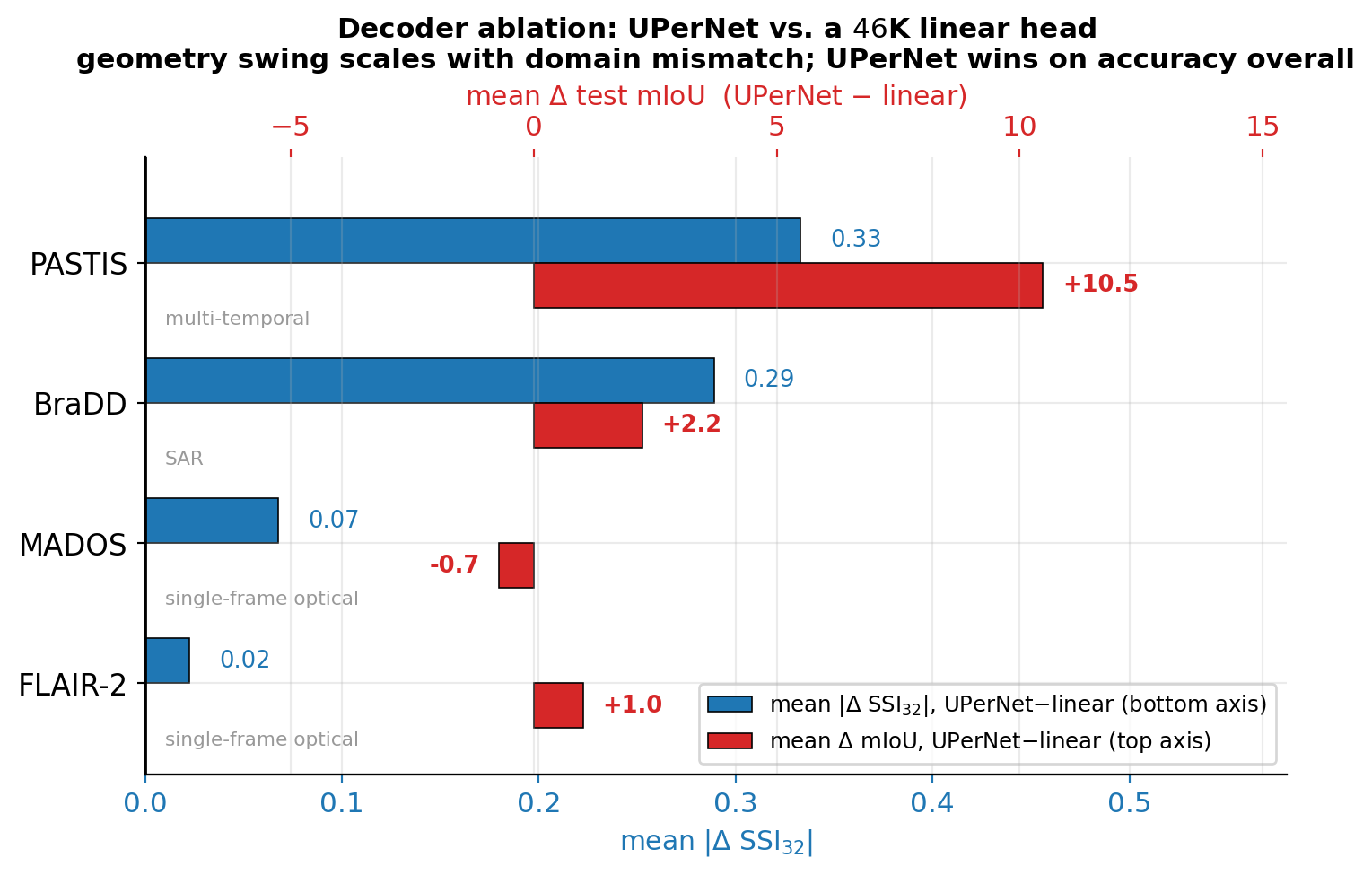}
\caption{
Decoder comparison across PASTIS, BraDD, MADOS, and FLAIR-2. Blue bars show the mean absolute change in $\mathrm{SSI}_{32}$; red bars show the mean mIoU difference between UPerNet and the linear decoder.
}
\label{fig:abl-decoder}
\end{figure}

We replace the linear decoder with UPerNet while keeping the encoder
architecture and training protocol unchanged (Fig.~\ref{fig:abl-decoder}).
UPerNet provides substantially greater task-specific
capacity~\citep{xiao2018unified}: its more expressive decoder with around 40M parameters can absorb
more of the downstream learning, which may reduce the need to reorganize
the pretrained encoder.

The effect of decoder choice is small on single-frame optical tasks (mean
$|\Delta\mathrm{SSI}_{32}|=0.02$--$0.07$) and much larger on the two
time-series tasks, PASTIS and BraDD ($0.29$--$0.33$), where UPerNet is
not only a much larger network but also uses a temporal attention encoder
to model the temporal dimension. On PASTIS, $\mathrm{SSI}_{32}$ increases
with UPerNet for every encoder, consistent with the decoder absorbing
part of the adaptation. UPerNet improves mean performance on PASTIS, BraDD, and FLAIR-2, and
slightly reduces it on MADOS. Decoder capacity thus influences both
downstream performance and the extent of encoder adaptation.

\subsection{Results on a Convolutional Encoder}
\label{sec:abl-cnn}

\begin{figure}[t]
\centering
\includegraphics[width=0.62\textwidth]{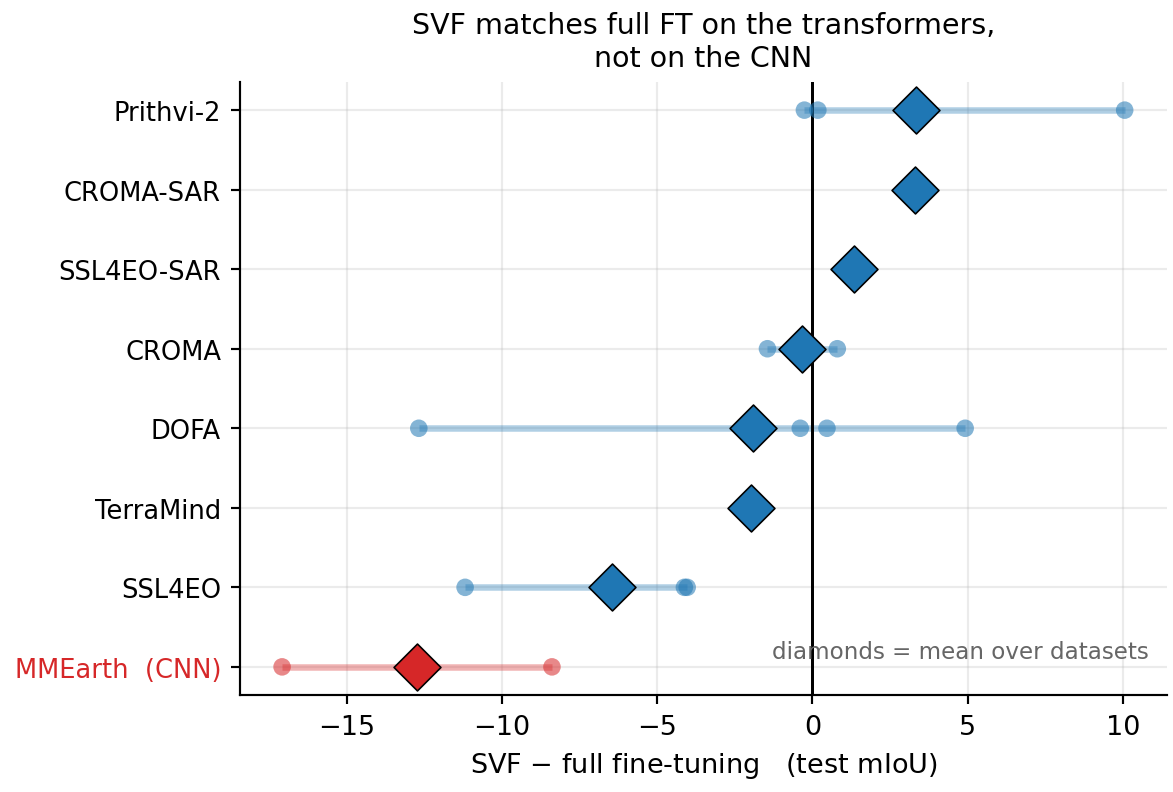}
\caption{
Difference in test mIoU between SVF and full fine-tuning across encoders. Diamonds indicate means
over datasets. MMEarth exhibits the largest mean performance deficit.
}
\label{fig:abl-cnn}
\end{figure}

To assess whether our observations extend beyond transformer encoders, we additionally evaluate MMEarth~\citep{nedungadi2024mmearth}, whose
encoder is based on the ConvNeXt family of convolutional architectures
\citep{liu2022convnet} (Fig.~\ref{fig:abl-cnn}). MMEarth exhibits the largest mean SVF deficit
among the evaluated models, with SVF trailing full fine-tuning by $12.7$ mIoU points. One possible contributing factor is the limited adaptation coverage of our SVF implementation, which modifies only
24 pointwise projection matrices in ConvNeXt.

Despite this performance gap, MMEarth retains high
$\mathrm{SSI}_{32}$ ($0.944$--$0.972$) while also exhibiting high ERA ($0.821$--$0.871$). This combination shows that strong preservation of
the dominant pretrained subspaces can coexist with updates distributed across many directions. Since this experiment includes only a single convolutional backbone, we cannot disentangle the effect of encoder
architecture from that of MMEarth's narrower layer dimensions.

\subsection{Null Baseline: Random Updates of the Same Norm}
\label{sec:null_baseline}

SSI and ERA could in principle be determined by the size of the update alone. To test this, we replace each update $\Delta W$ by a random matrix of the same norm, $\Delta W_G = G\,\|\Delta W\|_F / \|G\|_F$ with i.i.d.\ Gaussian $G$, and recompute SSI between $W_{\text{pre}}$ and $W_{\text{pre}} + \Delta W_G$ and ERA of $\Delta W_G$. By construction, $\rho$ is unchanged. We use the same matrices and aggregation as in the main results and average over 20 draws; the variation across draws is below 0.01.

Tab.~\ref{tab:null_baseline} reports the result for the TimeMatch runs of Tab.~\ref{tab:eo_timematch}. A random update of the same norm leaves the dominant subspaces far more intact than fine-tuning does: the null SSI averages 0.667/0.721 (attention/MLP), against 0.388/0.366 observed. The observed updates are also much more concentrated, with ERA of 0.612/0.687 against 0.876/0.968 for the null, for every backbone. Fine-tuning thus changes the dominant pretrained directions more than its norm alone would imply. The exception is Presto, whose observed SSI exceeds the null: its matrices are only 128 wide, so $k = 32$ covers a quarter of the space (Appendix~\ref{sec:prop_k}). SSI and $\rho$ remain correlated across models, since larger updates rotate the subspaces more, which is why we interpret the diagnostics jointly (Sec.~\ref{sec:joint_interpretation}).

\begin{table}[h]
\centering
\small
\begin{tabular}{@{}lcccccccc@{}}
\toprule
 & \multicolumn{2}{c}{SSI$_{32}$ observed} & \multicolumn{2}{c}{SSI$_{32}$ null} & \multicolumn{2}{c}{ERA observed} & \multicolumn{2}{c}{ERA null} \\
\cmidrule(lr){2-3}\cmidrule(lr){4-5}\cmidrule(lr){6-7}\cmidrule(lr){8-9}
Backbone & Attn & MLP & Attn & MLP & Attn & MLP & Attn & MLP \\
\midrule
CROMA      & .116 & .074 & .504 & .698 & .589 & .628 & .879 & .966 \\
DOFA       & .188 & .182 & .636 & .755 & .587 & .701 & .877 & .966 \\
Prithvi-2  & .327 & .175 & .822 & .770 & .597 & .707 & .876 & .966 \\
SSL4EO-MAE & .437 & .310 & .721 & .695 & .696 & .712 & .877 & .966 \\
AnySat     & .217 & .224 & .517 & .474 & .639 & .700 & .870 & .961 \\
Presto     & .715 & .759 & .595 & .703 & .416 & .519 & .878 & .966 \\
Tessera    & .716 & .836 & .871 & .954 & .760 & .840 & .875 & .984 \\
\midrule
Mean       & .388 & .366 & .667 & .721 & .612 & .687 & .876 & .968 \\
\bottomrule
\end{tabular}
\caption{Observed SSI$_{32}$ and ERA on TimeMatch (as in Tab.~\ref{tab:eo_timematch}) against a null in which each update is replaced by a Gaussian matrix of the same norm.}
\label{tab:null_baseline}
\end{table}

\section{Cross-Modal Transfer}
\label{sec:abl-xmodal}

We use CROMA's separate optical and SAR encoders to examine how
cross-modal transfer relates to preservation of pretrained structure.
Under full fine-tuning, we adapt the optical encoder to BraDD
and the SAR encoder to MADOS, comparing each with the encoder
pretrained on the target modality.

On BraDD, the optical encoder matches the SAR encoder at $54.1$
mIoU, but this performance parity accompanies substantial subspace
reorganization. Its $\mathrm{SSI}_{32}$ falls to approximately
$0.10$ in the early and middle blocks, recovering to
$0.47$--$0.55$ in the final blocks
(Fig.~\ref{fig:adaptation_and_pretraining}). This accentuates the
pattern observed in matched-modality experiments, where preservation
generally increases toward the output. The concentration of changes
in earlier layers is consistent with adaptation to unfamiliar
input statistics.

Conversely, on MADOS, the SAR encoder reaches only $40.7$ mIoU,
compared with $63.6$ for the optical encoder. Yet it preserves more
of its pretrained structure, with $\mathrm{SSI}_{32}$ ranging
from $0.66$ to $0.92$ across depth and a smaller relative weight
change ($\rho=0.28$ versus $0.46$). Greater preservation therefore
does not translate into better downstream performance in this case.

Together, these results reinforce the distinction between performance
and reuse: strong performance can coexist with extensive rewriting,
while high subspace preservation can accompany poor transfer.
Neither performance nor SSI alone establishes successful reuse
of the pretrained representation.

\section{Subspace Preservation Across Encoder Depth}
\label{app:depth}

\begin{figure}[!h]
\centering
\includegraphics[width=\textwidth]{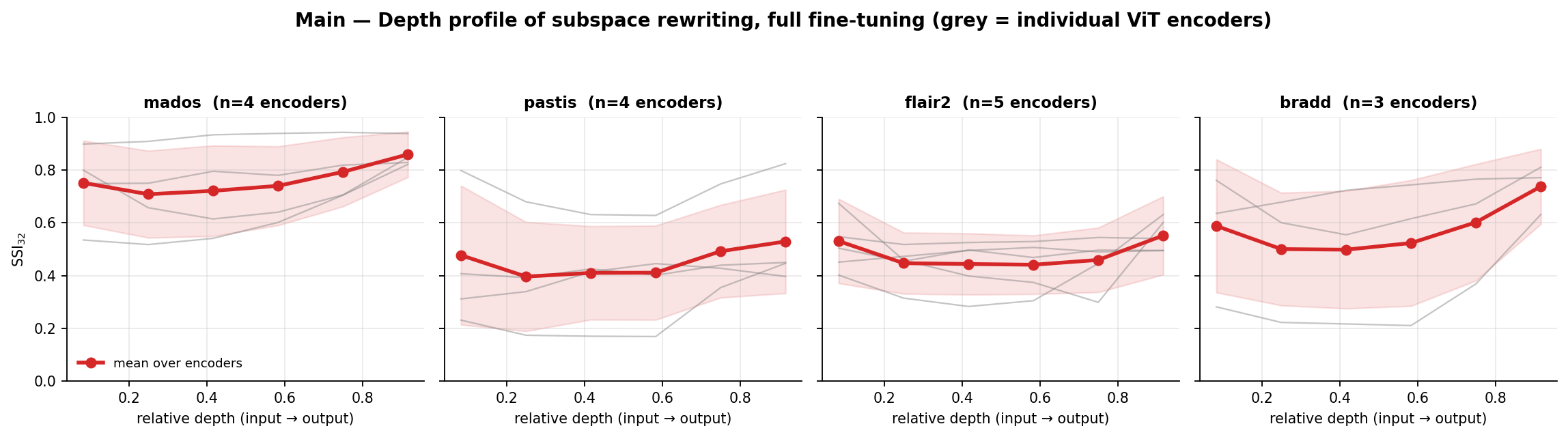}
\caption{Depth profile of subspace preservation under full fine-tuning, per benchmark.
Grey lines are individual encoders; red is the mean with $\pm 1$ std.}
\label{fig:depth-full}
\end{figure}

We normalize layer positions to $[0,1]$ and average
$\mathrm{SSI}_{32}$ over the attention and MLP matrices within
each block to compare encoders of different depths.
Across all four benchmarks and most individual encoders,
preservation is lowest in the early and middle layers and
increases toward the output (Fig.~\ref{fig:depth-full}).
The pooled mean decreases from $0.58$ in the first sixth
of the network to approximately $0.51$ through the early
and middle blocks, before rising to $0.66$ in the final sixth.
This pattern suggests that fine-tuning reorganizes the earlier
and intermediate transformations most strongly, while retaining
more of the later pretrained structure.

\section{Model Selection}
\label{app:selection}

\paragraph{Main EO encoders.}
Tab.~\ref{tab:panel} lists the five encoders evaluated in the main set of experiments. They are chosen to vary along the axes on which current geospatial foundation models actually differ, so that a systematic difference in adaptation
geometry cannot be attributed to a single shared design choice. The dominant
pretraining recipes in the current literature are all represented: masked
autoencoding in its canonical form (SSL4EO-MAE) and in a multitemporal variant
(Prithvi-2), cross-modal contrastive learning (CROMA), wavelength-conditioned
supervision (DOFA), and multimodal pretraining with a diffusion decoder (TerraMind). Optical-only,
SAR-capable and jointly optical-SAR encoders are all present, and CROMA additionally exposes a separate
encoder per modality. Only Prithvi-2 was pretrained with a temporal path, so the
multi-temporal benchmark contrasts a model built for it against four that were not.
Widths span $384$ to $1024$. All five are publicly released with pretrained weights and
span several model generations.

\paragraph{Additional encoders.} Four further backbones appear in specific
experiments rather than in the main sweep. \emph{MMEarth} (ConvNeXt, median width
$160$) is the convolutional encoder used in the
architecture ablation. \emph{Tessera} and \emph{Presto} (both
$128$-wide transformers with a native time-series path) and \emph{AnySat} are
evaluated on the crop-classification benchmark, where they provide the contrast
between encoders pretrained on time series and image ViTs routed through a pixel-set
path.

\begin{table}[h]
\centering
\small
\begin{tabular}{llcc}
\toprule
Encoder & Pretraining objective & Modality & $\min(m,n)$ \\
\midrule
CROMA      & cross-modal contrastive + MAE    & optical, SAR & 1024 \\
DOFA       & wavelength-conditioned           & multi-sensor & 768 \\
Prithvi-2  & masked autoencoding, multitemporal & optical    & 768 \\
TerraMind  & multimodal generative            & multi-sensor & 768 \\
SSL4EO-MAE & masked autoencoding              & optical, SAR & 384 \\
\bottomrule
\end{tabular}
\caption{Main EO encoders, their pretraining objectives, and input
modalities. Matrix dimensions are summarized by the median
$\min(m,n)$ across analyzed attention and MLP weight matrices.}

\label{tab:panel}
\end{table}

%

\section{Training Details}
\label{training_details}

We describe the dataset, encoder, adaptation, and
optimization settings below.

\subsection{Datasets and Splits}
\label{app:splits}

We use the published split of each benchmark and do not resample it
(Table~\ref{tab:splits}).

\begin{table}[h]
\centering
\small
\begin{tabular}{llrrrrl}
\toprule
Benchmark & Modality & Input & Bands & $T$ & Classes & Train / Val / Test \\
\midrule
MADOS     & optical S2        & $240^2$ & 11 & 1  & 15 & 1{,}433 / 642 / 728 \\
PASTIS & optical S2        & $128^2$ & 10 & 6  & 20 & folds 1-3 / 4 / 5 \\
FLAIR-2   & aerial RGB        & $512^2$ & \phantom{0}3 & 1  & 13 & 4{,}049 / 1{,}022 / 3{,}022 \\
BraDD     & SAR S1            & $48^2$  & \phantom{0}2 & 6  & \phantom{0}2 & official split \\
TimeMatch & optical pixel-set & 16 px   & 10 & 30 & \phantom{0}9 & cross-region \\
\bottomrule
\end{tabular}
\caption{Benchmarks. $T$ is the number of time steps after subsampling. FLAIR-2
is at 20\,cm ground sampling distance, the rest at 10\,m.}
\label{tab:splits}
\end{table}

\begin{itemize}
\item MADOS: 174 scenes, cropped. The 20\,m and 60\,m bands are resampled to
      10\,m by nearest neighbour. Label -1 (background) is ignored.
\item PASTIS: five official geographic folds. Each series is reduced to
      $T=6$ acquisitions at load time: series of at most 50 acquisitions
      are subsampled at evenly spaced indices in temporal order; longer
      series are first cut to a random, unordered subset of 50 (redrawn at
      every load), so their 6 frames are effectively a random draw without
      temporal order. This follows the PANGAEA loader. Class 19 (void) is
      ignored. The aerial and Sentinel-1 products are unused.
\item BraDD: VV and VH polarizations. The only SAR benchmark in the panel.
\item TimeMatch: 16 pixels per parcel, no spatial grid. We train on source
      tiles and test on a held-out tile.
\end{itemize}

\subsection{Model Configurations}
\label{app:model_config}

Each encoder runs at its released configuration and native resolution
(Table~\ref{tab:encoder_config}). We tap selected blocks per encoder
and pass them to the decoder as a feature pyramid.

\begin{table}[h]
\centering
\small
\begin{tabular}{llrrl}
\toprule
Encoder & Backbone & Input & Bands (opt./SAR) & Tapped blocks \\
\midrule
\multicolumn{5}{l}{\emph{Main EO encoders}} \\
CROMA      & ViT-L    & $120$ & 12 / 2 & 3, 5, 7, 11 \\
DOFA       & ViT-B    & $224$ & any$^\dagger$  & 3, 5, 7, 11 \\
Prithvi-2  & ViT-B    & $224$ & \phantom{0}6 / -- & 3, 5, 7, 11 \\
TerraMind  & ViT-B    & $224$ & 12 / 2 & 3, 5, 7, 11 \\
SSL4EO-MAE & ViT-S/16 & $224$ & 13 / 2 & 3, 5, 7, 11 \\
\midrule
\multicolumn{5}{l}{\emph{DINO-on-EO}} \\
DINOv2     & ViT-L/14 & $224$ & \phantom{0}3 & 5, 11, 17, 23 \\
DINOv3     & ViT-L/16 & $224$ & \phantom{0}3 & 5, 11, 17, 23 \\
\midrule
\multicolumn{5}{l}{\emph{Natural images}} \\
CLIP       & ViT-B/16 & $224$ & \phantom{0}3 & -- \\
DINOv2     & ViT-B/14 & $224$ & \phantom{0}3 & -- \\
DINOv3     & ViT-B/16 & $224$ & \phantom{0}3 & -- \\
\bottomrule
\end{tabular}
\caption{Encoder configurations. Bands is the number of input channels
expected (optical / SAR encoder). Natural-image backbones use their
standard classification or segmentation head. $^\dagger$DOFA is wavelength-conditioned and accepts the dataset's bands.}
\label{tab:encoder_config}
\end{table}

\begin{itemize}
\item \textbf{Band matching.} By name. Bands the encoder expects but the
      dataset lacks are zero-filled, bands it does not expect are dropped.
      MADOS lacks B9 and B10, PASTIS additionally lacks B1, so CROMA,
      TerraMind and SSL4EO-MAE receive one to three zero-filled bands.
      DOFA is wavelength-conditioned and takes the dataset's bands directly.
      The natural-image encoders take RGB only.
\item \textbf{Linear decoder} (main results). Concatenate the four tapped maps
      at native resolution, one $1\times1$ convolution to the class dimension,
      one bilinear upsample to label resolution. About 46\,K parameters on a
      768-wide encoder.
\item \textbf{UPerNet decoder} (ablation, Appendix~\ref{sec:abl-decoder}). 512
      channels, about 39\,M parameters. On multi-temporal benchmarks it merges
      time with an L-TAE module \citep{garnot2020ltae}, whereas the linear decoder stacks frames along
      the channel axis.
\item \textbf{TimeMatch} follows a pixel-set formulation
      \citep{garnot2020satellite}: each sampled pixel is processed
      independently by a shared encoder, and the resulting features
      are aggregated into a parcel representation for classification.
      We use a linear classification head throughout and exclude
      TimeMatch from the decoder ablation.
\item \textbf{Preprocessing.} MADOS is randomly cropped to the encoder input
      size in training, biased towards labelled regions, and evaluated by
      sliding window over the full tile. The others are resized and evaluated
      whole. All inputs are standardized per band with training-split
      statistics.
\item \textbf{DINO-on-EO experiments} use DINOv2-L/14 and DINOv3-L/16 with RGB inputs under the Earth-observation recipe: learning rate $10^{-4}$, 80 epochs, batch size 8 on MADOS and 4 on PASTIS. The DINOv2 vs.\ DINOv3 comparison in Fig.~\ref{fig:ssi_era_interpretation} uses the UPerNet decoder (with L-TAE on PASTIS), whereas the cross-domain comparison in Tab.~\ref{tab:cross_domain_rgb} uses the linear decoder, so the two sets of values differ. The natural-image experiments use DINOv2-Base and DINOv3-Base.
\end{itemize}

\subsection{Adaptation Settings}
\label{app:adaptation}

We compare three adaptation methods. Each is run from the pretrained
weights and from an architecture-matched random initialization.

\begin{itemize}
\item \textbf{Full fine-tuning.} All encoder and decoder parameters are
      trained.
\item \textbf{SVF.} Factorize each matrix once as $W = U\Sigma V^{\top}$,
      freeze $U$ and $V$, train only $\Sigma$. All singular values are trained,
      none truncated. Adaptation can rescale the pretrained directions but not
      rotate them.
\item \textbf{LoRA.} Freeze the matrix, add a trainable $BA$ of rank $r=8$
      scaled by $\alpha/r$ with $\alpha=16$. $A$ is initialized from a
      zero-mean normal with standard deviation $1/\sqrt{r}$, $B$ to zero, so
      the adapted matrix starts equal to the pretrained one. Rank and scaling
      are identical for Earth observation and natural images; only dropout on
      the low-rank branch differs (off for Earth observation, $0.05$ for
      natural images).
\end{itemize}

\paragraph{Scope of SVF and LoRA.} Both parameter-efficient methods wrap the
attention projections and feed-forward layers of every encoder block.
Normalization parameters, task heads and the patch-embedding projection are
excluded, and the decoder is always trained in full. The one exception is DOFA: the two feed-forward layers of its wavelength-conditioned
patch-embedding generator are wrapped as well, which is why DOFA has 50
adapted matrices in Table~\ref{tab:peft_budget} rather than 48. Neither
method reduces activation memory, only the number of updated parameters.

\paragraph{Random initialization.} The random control uses the same
architecture and skips loading the pretrained weights, so only the initial
weights differ. In both cases we save the encoder weights just before
adaptation, which gives an exact $W_{\mathrm{pre}}$ for the spectral
diagnostics.

\begin{table}[h]
\centering
\small
\begin{tabular}{lrrrr}
\toprule
Encoder & Encoder params & Adapted matrices & SVF trainable & LoRA trainable \\
\midrule
CROMA     & 303.0\,M & 96 & 98.3\,K (0.03\%) & 3.15\,M (1.04\%) \\
DOFA      & 111.3\,M & 50 & 37.1\,K (0.03\%) & 1.21\,M (1.09\%) \\
Prithvi-2 & \phantom{0}86.2\,M & 48 & 36.9\,K (0.04\%) & 1.18\,M (1.37\%) \\
TerraMind  & \phantom{0}87.3\,M & 61 & 46.8\,K (0.05\%) & 1.28\,M (1.47\%) \\
SSL4EO-MAE & \phantom{0}22.6\,M & 48 & 18.4\,K (0.08\%) & 0.59\,M (2.60\%) \\
\bottomrule
\end{tabular}
\caption{Trainable parameters, excluding the decoder. Percentages are relative
to the encoder.}
\label{tab:peft_budget}
\end{table}

\subsection{Optimization}
\label{app:optimization}

\paragraph{Earth observation.}
AdamW, learning rate $10^{-4}$, $\beta = (0.9, 0.999)$, weight decay $0.05$,
batch size 8 (4 for the two ViT-L DINO encoders on PASTIS, which process
each frame separately), seed 234. The learning rate drops by $10\times$ at
60\% and 90\% of total iterations. No warmup, no gradient clipping. All
models are trained for the full schedule, and both the reported scores and
the spectral diagnostics use the final checkpoint, since early stopping on
small validation sets would be noisy. The same learning rate is used for
all three methods and both initializations, with two exceptions: full
fine-tuning of Prithvi-2 on MADOS uses $10^{-5}$, because $10^{-4}$ was too
large, and LoRA and SVF for DOFA on PASTIS use $3\times10^{-4}$ and
$10^{-3}$, because $10^{-4}$ was too small. These are the only
learning-rate exceptions in
Tables~\ref{tab:eo_optical}--\ref{tab:eo_timematch}. Segmentation runs
80 epochs, except CROMA on MADOS at 160 and FLAIR-2 at 20, the latter
because its training split is about ten times larger. TimeMatch runs
30 epochs at batch size 32.


\paragraph{Natural images.}
10 epochs, batch size 64, $224\times224$, weight decay $10^{-4}$, seed 42. This
arm needs a few learning-rate exceptions (Table~\ref{tab:cv_lr}).

\begin{table}[h]
\centering
\small
\begin{tabular}{lll}
\toprule
Setting & Learning rate & Applies to \\
\midrule
Default          & $10^{-4}$        & everything not listed below \\
Full fine-tuning & $2\times10^{-5}$ & CIFAR-10 (all backbones), VOC (CLIP, DINOv2) \\
LoRA             & $5\times10^{-4}$ & VOC (DINOv3) \\
SVF & $5\times10^{-3}$ & VOC (DINOv3 and CLIP), Flowers-102 (CLIP) \\
\bottomrule
\end{tabular}
\caption{Learning rates for the natural-image experiments.
Appendix~\ref{app:lr_sensitivity} examines this sensitivity.}
\label{tab:cv_lr}
\end{table}


%% file: iclr2027_conference.bib
@article{corley2026no,
  title={No One Knows the State of the Art in Geospatial Foundation Models},
  author={Corley, Isaac and Lehmann, Nils and Robinson, Caleb and Tseng, Gabriel and Fuller, Anthony and Alemohammad, Hamed and Shelhamer, Evan and Marcus, Jennifer and Kerner, Hannah},
  journal={arXiv preprint arXiv:2605.12678},
  year={2026}
}

@article{marsocci2026pangaea,
  title   = {{PANGAEA}: Assessing Geospatial Foundation Models Capabilities
             through a Global and Inclusive Benchmark},
  author  = {Marsocci, Valerio and Jia, Yuru and Le Bellier, Georges
             and Kerekes, David and Zeng, Liang and Hafner, Sebastian
             and Gerard, Sebastian and Brune, Eric and Yadav, Ritu
             and Shibli, Ali and Fang, Heng and Ban, Yifang
             and Vergauwen, Maarten and Audebert, Nicolas
             and Nascetti, Andrea},
  journal = {IEEE Geoscience and Remote Sensing Magazine},
  volume  = {14},
  number  = {1},
  pages   = {245--285},
  year    = {2026},
  doi     = {10.1109/MGRS.2025.3628194}
}

@article{matsoukas2022makes,
  title={What makes transfer learning work for medical images: Feature reuse \& other factors},
  author={Matsoukas, Christos and Haslum, Johan Fredin and Sorkhei, Moein and S{\"o}derberg, Magnus and Smith, Kevin},
  journal={arXiv preprint arXiv:2203.01825},
  year={2022}
}

@article{brown2025alphaearth,
  title={Alphaearth foundations: An embedding field model for accurate and efficient global mapping from sparse label data},
  author={Brown, Christopher F and Kazmierski, Michal R and Pasquarella, Valerie J and Rucklidge, William J and Samsikova, Masha and Zhang, Chenhui and Shelhamer, Evan and Lahera, Estefania and Wiles, Olivia and Ilyushchenko, Simon and others},
  journal={arXiv preprint arXiv:2507.22291},
  year={2025}
}

@inproceedings{herzog2026olmoearth,
  title={Olmoearth: Stable latent image modeling for multimodal earth observation},
  author={Herzog, Henry and Bastani, Favyen and Zhang, Yawen and Tseng, Gabriel and Redmon, Joseph and Sablon, Hadrien and Park, Ryan and Morrison, Jacob and Buraczynski, Alexandra and Farley, Karen and others},
  booktitle={Proceedings of the IEEE/CVF Conference on Computer Vision and Pattern Recognition},
  pages={34806--34817},
  year={2026}
}

@inproceedings{kornblith2019similarity,
  title={Similarity of neural network representations revisited},
  author={Kornblith, Simon and Norouzi, Mohammad and Lee, Honglak and Hinton, Geoffrey},
  booktitle={International conference on machine learning},
  pages={3519--3529},
  year={2019},
  organization={PMlR}
}

@inproceedings{garnot2020satellite,
  title={Satellite image time series classification with pixel-set encoders and temporal self-attention},
  author    = {Sainte Fare Garnot, Vivien and Landrieu, Loic
               and Giordano, Sebastien and Chehata, Nesrine},
  booktitle = {Proceedings of the IEEE/CVF Conference on Computer
               Vision and Pattern Recognition (CVPR)},  pages={12322--12331},
  year={2020}
}

@article{bommasani2021foundation,
  title   = {On the Opportunities and Risks of Foundation Models},
  author  = {Bommasani, Rishi and Hudson, Drew A. and Adeli, Ehsan and others},
  journal = {arXiv preprint arXiv:2108.07258},
  year    = {2021}
}

@inproceedings{brown2020language,
  title     = {Language Models are Few-Shot Learners},
  author    = {Brown, Tom B. and Mann, Benjamin and Ryder, Nick and others},
  booktitle = {Advances in Neural Information Processing Systems},
  volume    = {33},
  year      = {2020}
}

@article{oquab2023dinov2,
  title   = {{DINOv2}: Learning Robust Visual Features without Supervision},
  author  = {Oquab, Maxime and Darcet, Timoth{\'e}e and Moutakanni, Th{\'e}o and others},
  journal = {Transactions on Machine Learning Research},
  year    = {2024}
}

@inproceedings{cong2022satmae,
  title     = {{SatMAE}: Pre-training Transformers for Temporal and Multi-Spectral Satellite Imagery},
  author    = {Cong, Yezhen and Khanna, Samar and Meng, Chenlin and Liu, Patrick
               and Rozi, Erik and He, Yutong and Burke, Marshall
               and Lobell, David B. and Ermon, Stefano},
  booktitle = {Advances in Neural Information Processing Systems},
  volume    = {35},
  year      = {2022}
}

@inproceedings{fuller2023croma,
  title     = {{CROMA}: Remote Sensing Representations with Contrastive Radar-Optical Masked Autoencoders},
  author    = {Fuller, Anthony and Millard, Koreen and Green, James R.},
  booktitle = {Advances in Neural Information Processing Systems},
  volume    = {36},
  year      = {2023}
}

@article{szwarcman2025prithvi,
  title   = {{Prithvi-EO-2.0}: A Versatile Multitemporal Foundation Model for {E}arth Observation Applications},
  author  = {Szwarcman, Daniela and Roy, Sujit and Fraccaro, Paolo and others},
  journal = {IEEE Transactions on Geoscience and Remote Sensing},
  volume  = {64},
  pages   = {1--20},
  doi     = {10.1109/TGRS.2025.3642610},
  year    = {2025}
}

@article{bodnar2025aurora,
  title   = {A Foundation Model for the {E}arth System},
  author  = {Bodnar, Cristian and Bruinsma, Wessel P. and Lucic, Ana and others},
  journal = {Nature},
  volume  = {641},
  pages   = {1180--1187},
  doi     = {10.1038/s41586-025-09005-y},
  year    = {2025}
}

@inproceedings{hu2022lora,
  title     = {{LoRA}: Low-Rank Adaptation of Large Language Models},
  author    = {Hu, Edward J. and Shen, Yelong and Wallis, Phillip
               and Allen-Zhu, Zeyuan and Li, Yuanzhi and Wang, Shean
               and Wang, Lu and Chen, Weizhu},
  booktitle = {International Conference on Learning Representations},
  year      = {2022}
}

@inproceedings{sun2022svf,
  title     = {Singular Value Fine-tuning: Few-shot Segmentation
               Requires Few-parameters Fine-tuning},
  author    = {Sun, Yanpeng and Chen, Qiang and He, Xiangyu
               and Wang, Jian and Feng, Haocheng and Han, Junyu
               and Ding, Errui and Cheng, Jian and Li, Zechao
               and Wang, Jingdong},
  booktitle = {Advances in Neural Information Processing Systems},
  volume    = {35},
  year      = {2022},
  doi       = {10.52202/068431-2717}
}

@inproceedings{zhang2024spectraladapter,
  title     = {Spectral Adapter: Fine-Tuning in Spectral Space},
  author    = {Zhang, Fangzhao and Pilanci, Mert},
  booktitle = {Advances in Neural Information Processing Systems},
  volume    = {37},
  year      = {2024},
  doi       = {10.52202/079017-4158}
}

@inproceedings{zhao2025sst,
  title     = {Sparse Spectral Training and Inference on {E}uclidean
               and Hyperbolic Neural Networks},
  author    = {Zhao, Jialin and Zhang, Yingtao and Li, Xinghang
               and Liu, Huaping and Cannistraci, Carlo Vittorio},
  booktitle = {Proceedings of the 42nd International Conference
               on Machine Learning},
  series    = {Proceedings of Machine Learning Research},
  volume    = {267},
  pages     = {77976--78002},
  publisher = {PMLR},
  year      = {2025}
}

@inproceedings{borodin2026diffract,
  title     = {Diffract: Spectral View of {LLM} Domain Adaptation},
  author    = {Borodin, Nikita and Krylova, Maria
               and Zabolotnyi, Artem and Aspisov, Dmitry
               and Shikov, Egor and Tyuplyaev, Nikita
               and Travkin, Oleg and Alferov, Roman
               and Vinichenko, Dmitry},
  booktitle = {Proceedings of the 43rd International Conference on Machine Learning},
  year      = {2026},
  url       = {https://openreview.net/forum?id=XBUHoiAGDE}
}

@article{lam2023graphcast,
  title   = {Learning Skillful Medium-Range Global Weather Forecasting},
  author  = {Lam, Remi and Sanchez-Gonzalez, Alvaro
             and Willson, Matthew and Wirnsberger, Peter
             and Fortunato, Meire and Alet, Ferran
             and Ravuri, Suman and Ewalds, Timo
             and Eaton-Rosen, Zach and Hu, Weihua
             and Merose, Alexander and Hoyer, Stephan
             and Holland, George and Vinyals, Oriol
             and Stott, Jacklynn and Pritzel, Alexander
             and Mohamed, Shakir and Battaglia, Peter},
  journal = {Science},
  volume  = {382},
  number  = {6677},
  pages   = {1416--1421},
  year    = {2023},
  doi     = {10.1126/science.adi2336}
}

@article{elhage2022superposition,
  title   = {Toy Models of Superposition},
  author  = {Elhage, Nelson and Hume, Tristan and Olsson, Catherine
             and Schiefer, Nicholas and Henighan, Tom and Kravec, Shauna
             and Hatfield-Dodds, Zac and Lasenby, Robert and Drain, Dawn
             and Chen, Carol and Grosse, Roger and McCandlish, Sam
             and Kaplan, Jared and Amodei, Dario and Wattenberg, Martin
             and Olah, Christopher},
  journal = {Transformer Circuits Thread},
  year    = {2022},
  url     = {https://transformer-circuits.pub/2022/toy_model/}
}

@misc{millidge2022svd,
  title        = {The Singular Value Decompositions of Transformer Weight Matrices
                  are Highly Interpretable},
  author       = {Millidge, Beren and Black, Sid},
  year         = {2022},
  howpublished = {AI Alignment Forum},
  url          = {https://www.alignmentforum.org/posts/mkbGjzxD8d8XqKHzA/the-singular-value-decompositions-of-transformer-weight}
}

@inproceedings{staats2025small,
  title     = {Small Singular Values Matter:
               A Random Matrix Analysis of Transformer Models},
  author    = {Staats, Max and Thamm, Matthias and Rosenow, Bernd},
  booktitle = {Advances in Neural Information Processing Systems},
  volume    = {38},
  year      = {2025}
}

@inproceedings{meng2022rome,
  title     = {Locating and Editing Factual Associations in {GPT}},
  author    = {Meng, Kevin and Bau, David and Andonian, Alex
               and Belinkov, Yonatan},
  booktitle = {Advances in Neural Information Processing Systems},
  volume    = {35},
  year      = {2022}
}

@inproceedings{aghajanyan2021intrinsic,
  title     = {Intrinsic Dimensionality Explains the Effectiveness
               of Language Model Fine-Tuning},
  author    = {Aghajanyan, Armen and Gupta, Sonal and Zettlemoyer, Luke},
  booktitle = {Proceedings of the 59th Annual Meeting of the Association
               for Computational Linguistics and the 11th International
               Joint Conference on Natural Language Processing
               (Volume 1: Long Papers)},
  pages     = {7319--7328},
  publisher = {Association for Computational Linguistics},
  year      = {2021},
  doi       = {10.18653/v1/2021.acl-long.568}
}

@inproceedings{turkoglu2026sve,
  title     = {Quantifying the Uncertainty of Foundation Models
               with Singular Value Ensembles},
  author    = {Turkoglu, Mehmet Ozgur and M{\"u}hlematter, Dominik J.
               and Becker, Alexander and Schindler, Konrad
               and Aasen, Helge},
  booktitle = {Proceedings of the 43rd International Conference
               on Machine Learning},
  year      = {2026}
}

@article{kikaki2024mados,
  title   = {Detecting Marine Pollutants and Sea Surface Features
             with Deep Learning in {Sentinel-2} Imagery},
  author  = {Kikaki, Katerina and Kakogeorgiou, Ioannis
             and Hoteit, Ibrahim and Karantzalos, Konstantinos},
  journal = {ISPRS Journal of Photogrammetry and Remote Sensing},
  volume  = {210},
  pages   = {39--54},
  year    = {2024},
  doi     = {10.1016/j.isprsjprs.2024.02.017}
}

@article{garioud2023flair2,
  title         = {{FLAIR} \#2: Textural and Temporal Information for
                   Semantic Segmentation from Multi-Source Optical Imagery},
  author        = {Garioud, Anatol and De Wit, Apolline and Poup{\'e}e, Marc
                   and Valette, Marion and Giordano, S{\'e}bastien
                   and Wattrelos, Boris},
  year          = {2023},
  journal       = {arXiv preprint arXiv:2305.14467}
}

@inproceedings{garnot2021pastis,
  title     = {Panoptic Segmentation of Satellite Image Time Series with
               Convolutional Temporal Attention Networks},
  author    = {Sainte Fare Garnot, Vivien and Landrieu, Loic},
  booktitle = {Proceedings of the IEEE/CVF International Conference
               on Computer Vision (ICCV)},
  pages     = {4852--4861},
  year      = {2021},
  doi       = {10.1109/ICCV48922.2021.00483}
}

@article{nyborg2022timematch,
  title   = {{TimeMatch}: Unsupervised Cross-Region Adaptation by
             Temporal Shift Estimation},
  author  = {Nyborg, Joachim and Pelletier, Charlotte
             and Lef{\`e}vre, S{\'e}bastien and Assent, Ira},
  journal = {ISPRS Journal of Photogrammetry and Remote Sensing},
  volume  = {188},
  pages   = {301--313},
  year    = {2022},
  doi     = {10.1016/j.isprsjprs.2022.04.018}
}

@article{karaman2023bradd,
  title   = {Deforestation Detection in the {Amazon} with {Sentinel-1}
             {SAR} Image Time Series},
  author  = {Karaman, Kaan and Sainte Fare Garnot, Vivien
             and Wegner, Jan Dirk},
  journal = {ISPRS Annals of the Photogrammetry, Remote Sensing
             and Spatial Information Sciences},
  volume  = {X-1/W1-2023},
  pages   = {835--842},
  year    = {2023},
  doi     = {10.5194/isprs-annals-X-1-W1-2023-835-2023}
}

@article{xiong2024dofa,
  title         = {Neural Plasticity-Inspired Multimodal Foundation Model
                   for {E}arth Observation},
  author        = {Xiong, Zhitong and Wang, Yi and Zhang, Fahong
                   and Stewart, Adam J. and Hanna, Jo{\"e}lle
                   and Borth, Damian and Papoutsis, Ioannis
                   and Le Saux, Bertrand and Camps-Valls, Gustau
                   and Zhu, Xiao Xiang},
  year          = {2024},
  journal       = {arXiv preprint arXiv:2403.15356}
}

@inproceedings{jakubik2025terramind,
  title     = {{TerraMind}: Large-Scale Generative Multimodality
               for {E}arth Observation},
  author    = {Jakubik, Johannes and Yang, Felix and Blumenstiel, Benedikt
               and Scheurer, Erik and Sedona, Rocco
               and Maurogiovanni, Stefano and Bosmans, Jente
               and Dionelis, Nikolaos and Marsocci, Valerio
               and Kopp, Niklas and Ramachandran, Rahul
               and Fraccaro, Paolo and Brunschwiler, Thomas
               and Cavallaro, Gabriele and Bernabe-Moreno, Juan
               and Long{\'e}p{\'e}, Nicolas},
  booktitle = {Proceedings of the IEEE/CVF International Conference
               on Computer Vision (ICCV)},
  pages     = {7383--7394},
  year      = {2025},
  doi       = {10.1109/ICCV51701.2025.00693}
}

@article{wang2023ssl4eo,
  title   = {{SSL4EO-S12}: A Large-Scale Multimodal, Multitemporal Dataset
             for Self-Supervised Learning in {E}arth Observation},
  author  = {Wang, Yi and Braham, Nassim Ait Ali and Xiong, Zhitong
             and Liu, Chenying and Albrecht, Conrad M. and Zhu, Xiao Xiang},
  journal = {IEEE Geoscience and Remote Sensing Magazine},
  volume  = {11},
  number  = {3},
  pages   = {98--106},
  year    = {2023},
  doi     = {10.1109/MGRS.2023.3281651}
}

@inproceedings{radford2021clip,
  title     = {Learning Transferable Visual Models From Natural Language
               Supervision},
  author    = {Radford, Alec and Kim, Jong Wook and Hallacy, Chris
               and Ramesh, Aditya and Goh, Gabriel and Agarwal, Sandhini
               and Sastry, Girish and Askell, Amanda and Mishkin, Pamela
               and Clark, Jack and Krueger, Gretchen and Sutskever, Ilya},
  booktitle = {Proceedings of the 38th International Conference on
               Machine Learning},
  series    = {Proceedings of Machine Learning Research},
  volume    = {139},
  pages     = {8748--8763},
  publisher = {PMLR},
  year      = {2021},
  url       = {https://proceedings.mlr.press/v139/radford21a.html}
}

@article{simeoni2025dinov3,
  title         = {{DINOv3}},
  author        = {Sim{\'e}oni, Oriane and Vo, Huy V. and Seitzer, Maximilian
                   and Baldassarre, Federico and Oquab, Maxime and Jose, Cijo
                   and Khalidov, Vasil and Szafraniec, Marc and Yi, Seungeun
                   and Ramamonjisoa, Micha{\"e}l and Massa, Francisco
                   and Haziza, Daniel and Wehrstedt, Luca and Wang, Jianyuan
                   and Darcet, Timoth{\'e}e and Moutakanni, Th{\'e}o
                   and Sentana, Leonel and Roberts, Claire and Vedaldi, Andrea
                   and Tolan, Jamie and Brandt, John and Couprie, Camille
                   and Mairal, Julien and J{\'e}gou, Herv{\'e}
                   and Labatut, Patrick and Bojanowski, Piotr},
  year          = {2025},
  journal       = {arXiv preprint arXiv:2508.10104}
}

@inproceedings{nilsback2008flowers,
  title     = {Automated Flower Classification over a Large Number of Classes},
  author    = {Nilsback, Maria-Elena and Zisserman, Andrew},
  booktitle = {2008 Sixth Indian Conference on Computer Vision,
               Graphics \& Image Processing},
  pages     = {722--729},
  publisher = {IEEE},
  year      = {2008},
  doi       = {10.1109/ICVGIP.2008.47}
}

@techreport{krizhevsky2009cifar,
  title       = {Learning Multiple Layers of Features from Tiny Images},
  author      = {Krizhevsky, Alex},
  institution = {University of Toronto},
  year        = {2009},
  url         = {https://www.cs.toronto.edu/~kriz/learning-features-2009-TR.pdf}
}

@inproceedings{parkhi2012pets,
  title     = {Cats and Dogs},
  author    = {Parkhi, Omkar M. and Vedaldi, Andrea and Zisserman, Andrew
               and Jawahar, C. V.},
  booktitle = {2012 IEEE Conference on Computer Vision and
               Pattern Recognition (CVPR)},
  pages     = {3498--3505},
  publisher = {IEEE},
  year      = {2012},
  doi       = {10.1109/CVPR.2012.6248092}
}

@article{everingham2010voc,
  title   = {The {Pascal} Visual Object Classes ({VOC}) Challenge},
  author  = {Everingham, Mark and Van Gool, Luc
             and Williams, Christopher K. I. and Winn, John
             and Zisserman, Andrew},
  journal = {International Journal of Computer Vision},
  volume  = {88},
  number  = {2},
  pages   = {303--338},
  year    = {2010},
  doi     = {10.1007/s11263-009-0275-4}
}

@article{bjorck1973numerical,
  title={Numerical methods for computing angles between linear subspaces},
  author={Bj{\"o}rck, {\AA}ke and Golub, Gene H.},
  journal={Mathematics of Computation},
  volume={27},
  number={123},
  pages={579--594},
  year={1973}
}

@inproceedings{roy2007effective,
  author    = {Roy, Olivier and Vetterli, Martin},
  title     = {The Effective Rank: A Measure of Effective Dimensionality},
  booktitle = {2007 15th European Signal Processing Conference},
  year      = {2007},
  pages     = {606--610}
}

@article{si2025weight,
  title   = {Weight Spectra Induced Efficient Model Adaptation},
  author  = {Si, Chongjie and Yang, Xuankun and Liu, Muqing and Wang, Yadao and Yang, Xiaokang and Su, Wenbo and Zheng, Bo and Shen, Wei},
  journal = {arXiv preprint arXiv:2505.23099},
  year    = {2025}
}

@inproceedings{neyshabur2020being,
  title     = {What Is Being Transferred in Transfer Learning?},
  author    = {Neyshabur, Behnam and Sedghi, Hanie and Zhang, Chiyuan},
  booktitle = {Advances in Neural Information Processing Systems},
  volume    = {33},
  year      = {2020}
}

@inproceedings{shuttleworth2025lora,
  title     = {{LoRA} vs Full Fine-tuning: An Illusion of Equivalence},
  author    = {Shuttleworth, Reece and Andreas, Jacob and Torralba, Antonio and Sharma, Pratyusha},
  booktitle = {Advances in Neural Information Processing Systems},
  volume    = {38},
  year      = {2025}
}

@inproceedings{xiao2018unified,
  title     = {Unified Perceptual Parsing for Scene Understanding},
  author    = {Xiao, Tete and Liu, Yingcheng and Zhou, Bolei and Jiang, Yuning and Sun, Jian},
  booktitle = {Proceedings of the European Conference on Computer Vision (ECCV)},
  year      = {2018},
  pages     = {432--448},
}

@inproceedings{nedungadi2024mmearth,
  title     = {{MMEarth}: Exploring Multi-Modal Pretext Tasks for Geospatial Representation Learning},
  author    = {Nedungadi, Vishal and Kariryaa, Ankit and Oehmcke, Stefan and Belongie, Serge and Igel, Christian and Lang, Nico},
  booktitle = {Proceedings of the European Conference on Computer Vision (ECCV)},
  pages     = {164--182},
  year      = {2024},
  doi       = {10.1007/978-3-031-73039-9_10}
}

@inproceedings{liu2022convnet,
  title     = {A {ConvNet} for the 2020s},
  author    = {Liu, Zhuang and Mao, Hanzi and Wu, Chao-Yuan and Feichtenhofer, Christoph and Darrell, Trevor and Xie, Saining},
  booktitle = {Proceedings of the IEEE/CVF Conference on Computer Vision and Pattern Recognition (CVPR)},
  pages     = {11976--11986},
  year      = {2022},
  doi       = {10.1109/CVPR52688.2022.01167}
}

@inproceedings{bonafilia2020sen1floods11,
  title     = {{Sen1Floods11}: A Georeferenced Dataset to Train and Test
               Deep Learning Flood Algorithms for {Sentinel-1}},
  author    = {Bonafilia, Derrick and Tellman, Beth and Anderson, Tyler
               and Issenberg, Erica},
  booktitle = {Proceedings of the IEEE/CVF Conference on Computer
               Vision and Pattern Recognition Workshops},
  pages     = {835--845},
  year      = {2020},
  doi       = {10.1109/CVPRW50498.2020.00113}
}

@misc{phillips2023hlsburnscars,
  title        = {{HLS} Foundation Burnscars Dataset},
  author       = {Phillips, Christopher and Roy, Sujit and Ankur, Kumar
                  and Ramachandran, Rahul},
  year         = {2023},
  howpublished = {Hugging Face dataset},
  doi          = {10.57967/hf/0956},
  url          = {https://huggingface.co/datasets/ibm-nasa-geospatial/hls_burn_scars}
}

@inproceedings{feng2026tessera,
  title     = {{TESSERA}: Temporal Embeddings of Surface Spectra
               for {E}arth Representation and Analysis},
  author    = {Feng, Zhengpeng and Atzberger, Clement and Jaffer, Sadiq
               and Knezevic, Jovana and Sormunen, Silja and Young, Robin
               and Lisaius, Madeline C. and Immitzer, Markus
               and Jackson, Toby and Ball, James and Coomes, David A.
               and Madhavapeddy, Anil and Blake, Andrew
               and Keshav, Srinivasan},
  booktitle = {Proceedings of the IEEE/CVF Conference on Computer
               Vision and Pattern Recognition (CVPR)},
  pages     = {34818--34831},
  year      = {2026}
}

@article{tseng2023presto,
  title   = {Lightweight, Pre-trained Transformers for Remote Sensing
             Timeseries},
  author  = {Tseng, Gabriel and Cartuyvels, Ruben and Zvonkov, Ivan
             and Purohit, Mirali and Rolnick, David and Kerner, Hannah},
  journal = {arXiv preprint arXiv:2304.14065},
  year    = {2023}
}

@inproceedings{astruc2025anysat,
  title     = {{AnySat}: One {E}arth Observation Model for Many
               Resolutions, Scales, and Modalities},
  author    = {Astruc, Guillaume and Gonthier, Nicolas
               and Mallet, Cl{\'e}ment and Landrieu, Loic},
  booktitle = {Proceedings of the IEEE/CVF Conference on Computer
               Vision and Pattern Recognition (CVPR)},
  pages     = {19530--19540},
  year      = {2025}
}

@inproceedings{garnot2020ltae,
  title     = {Lightweight Temporal Self-Attention for Classifying
               Satellite Images Time Series},
  author    = {Sainte Fare Garnot, Vivien and Landrieu, Loic},
  booktitle = {Advanced Analytics and Learning on Temporal Data
               (AALTD 2020)},
  series    = {Lecture Notes in Computer Science},
  volume    = {12588},
  pages     = {171--181},
  publisher = {Springer},
  year      = {2020},
  doi       = {10.1007/978-3-030-65742-0_12}
}

@inproceedings{liu2025lift,
  title     = {{LIFT} the Veil for the Truth: Principal Weights Emerge after Rank
               Reduction for Reasoning-Focused Supervised Fine-Tuning},
  author    = {Liu, Zihang and Pang, Tianyu and Balabanov, Oleg and Yang, Chaoqun
               and Huang, Tianjin and Yin, Lu and Yang, Yaoqing and Liu, Shiwei},
  booktitle = {Proceedings of the 42nd International Conference on Machine Learning},
  series    = {Proceedings of Machine Learning Research},
  volume    = {267},
  year      = {2025}
}

@article{biderman2024lora,
  title   = {{LoRA} Learns Less and Forgets Less},
  author  = {Biderman, Dan and Portes, Jacob and Gonzalez Ortiz, Jose Javier and
             Paul, Mansheej and Greengard, Philip and Jennings, Connor and King,
             Daniel and Havens, Sam and Chiley, Vitaliy and Frankle, Jonathan and
             Blakeney, Cody and Cunningham, John P.},
  journal = {Transactions on Machine Learning Research},
  year    = {2024}
}

@inproceedings{kumar2022finetuning,
  title     = {Fine-Tuning can Distort Pretrained Features and Underperform
               Out-of-Distribution},
  author    = {Kumar, Ananya and Raghunathan, Aditi and Jones, Robbie and Ma, Tengyu
               and Liang, Percy},
  booktitle = {International Conference on Learning Representations},
  year      = {2022}
}

@inproceedings{raghu2019transfusion,
  title     = {Transfusion: Understanding Transfer Learning for Medical Imaging},
  author    = {Raghu, Maithra and Zhang, Chiyuan and Kleinberg, Jon and Bengio, Samy},
  booktitle = {Advances in Neural Information Processing Systems},
  volume    = {32},
  year      = {2019}
}

@inproceedings{merchant2020happens,
  title     = {What Happens To {BERT} Embeddings During Fine-tuning?},
  author    = {Merchant, Amil and Rahimtoroghi, Elahe and Pavlick, Ellie and
               Tenney, Ian},
  booktitle = {Proceedings of the Third BlackboxNLP Workshop on Analyzing and
               Interpreting Neural Networks for NLP},
  year      = {2020}
}

@inproceedings{sosa2024effective,
  title     = {How Effective is Pre-training of Large Masked Autoencoders
               for Downstream {E}arth Observation Tasks?},
  author    = {Sosa, Jose and Aloulou, Mohamed and Rukhovich, Danila
               and Sleimi, Rim and Changaival, Boonyarit and Kacem, Anis
               and Aouada, Djamila},
  booktitle = {British Machine Vision Conference (BMVC) Workshops, MVEO},
  year      = {2024},
  url       = {https://bmva-archive.org.uk/bmvc/2024/workshops/MVEO/paper5.pdf}
}
